\documentclass[lettersize,journal]{IEEEtran}
\usepackage{amsmath,amsfonts}
\usepackage[ruled]{algorithm2e}  
\usepackage{array}
\usepackage[caption=false,font=normalsize,labelfont=sf,textfont=sf]{subfig}
\usepackage{textcomp}
\usepackage{stfloats}
\usepackage{url}
\usepackage{verbatim}
\usepackage{graphicx}
\usepackage{threeparttable}
\usepackage{cite}
\usepackage{color}
\usepackage[table,xcdraw]{xcolor}
\usepackage{multirow}

\begin{document}

\title{FedTP: Federated Learning by Transformer Personalization}

\author{Hongxia Li, Zhongyi Cai, Jingya Wang, Jiangnan Tang, Weiping Ding, Chin-Teng Lin, and Ye Shi
\thanks{This work was supported by the Shanghai Sailing Program
(22YF1428800, 21YF1429400), Shanghai Local college
capacity building program (23010503100), and Shanghai Frontiers Science Center of Humancentered Artificial Intelligence (ShangHAI). This work was also supported in part by the National Natural Science Foundation of China under Grant 61976120, in part by the Natural Science Key Foundation of Jiangsu Education Department under Grant 21KJA510004, in part by the
Australian Research Council (ARC) under Discovery Grant DP210101093
and Discovery Grant DP220100803. (Corresponding authors: Ye Shi)} 
\thanks{Hongxia Li and Zhongyi Cai contributed equally to this work. Hongxia Li, Zhongyi Cai, Jingya Wang, Jiangnan Tang and Ye Shi are with the School of Information Science and Technology, ShanghaiTech University, Shanghai 201210, China (e-mail:lihx2@shanghaitech.edu.cn, caizhy@shanghaitech.edu.cn, wangjingya@shanghaitech.edu.cn, tangjn2022@shanghaitech.edu.cn, shiye@shanghaitech.edu.cn).}
\thanks{Weiping Ding is with School of Information Science and Technology, Nantong University, Nantong 226019, China (e-mail: dwp9988@163.com)}
\thanks{Chin-Teng Lin is with the School of Computer Science, University of Technology Sydney, Broadway, NSW 2007, Australia (e-mail: chin-teng.lin@uts.edu.au).}}

\maketitle

\begin{abstract}
Federated learning is an emerging learning paradigm where multiple clients collaboratively train a machine learning model in a privacy-preserving manner. Personalized federated learning extends this paradigm to overcome heterogeneity across clients by learning personalized models. Recently, there have been some initial attempts to apply Transformers to federated learning. However, the impacts of federated learning algorithms on self-attention have not yet been studied. In this paper, we investigate this relationship and reveals that federated averaging algorithms actually have a negative impact on self-attention in cases of data heterogeneity, which limits the capabilities of the Transformer model in federated learning settings. To address this issue, we propose FedTP, a novel Transformer-based federated learning framework that learns personalized self-attention for each client while aggregating the other parameters among the clients. Instead of using a vanilla personalization mechanism that maintains personalized self-attention layers of each client locally, we develop a \emph{learn-to-personalize} mechanism to further encourage the cooperation among clients and to increase the scalability and generalization of FedTP. Specifically, we achieve this by learning a hypernetwork on the server that outputs the personalized projection matrices of self-attention layers to generate client-wise queries, keys and values. Furthermore, we present the generalization bound for FedTP with the \emph{learn-to-personalize} mechanism. 
Extensive experiments verify that FedTP with the \emph{learn-to-personalize} mechanism yields state-of-the-art performance in non-IID scenarios. Our code is available online\footnote{\url{https://github.com/zhyczy/FedTP}}. 
\end{abstract}


\section{Introduction}
Federated learning \cite{mcmahan2017communication} is a framework that learns a shared global model from multiple disjointed clients without sharing their own data. In federated learning, each client trains a model using its own local data and only sends model updates back to the server. In this way, federated learning overcomes a range of problems with both data privacy and communications overheads. However, in situations where there is data heterogeneity and system heterogeneity across the clients, learning a single global model may fail. Accordingly, personalized federated learning has emerged as an extension to federated learning to overcome these challenges. The paradigm works by learning personalized models instead of a single global model, while still benefiting from joint training. 

\begin{figure}[t]
\centering
\includegraphics[width=0.98\columnwidth]{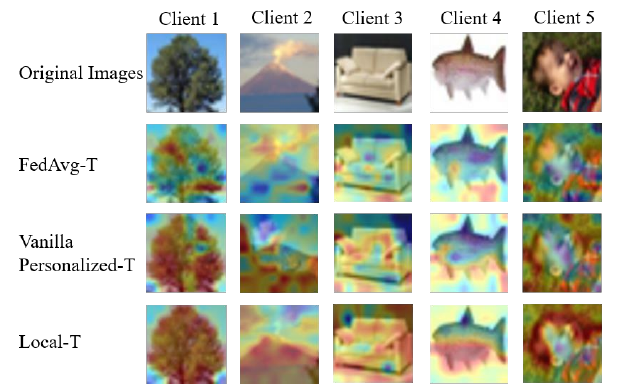} 
\caption{An example showing the attention maps of Local-T, FedAvg-T and Vanilla Personalized-T over 5 clients with different local datasets. These results show that self-attention either trained locally or by a Vanilla personalized-T model can focus on task-specific information well. However, FedAvg-T disturbs these information.}
\label{attention_map}
\end{figure}

Data heterogeneity is a common problem in the real world, since the training data collected from different clients varies heavily both in size and distribution \cite{sattler2019robust,sattler2020clustered,shi2020consensus,zhang2020hierarchical,shi2021distributed}. 
Numerous methods have been developed for federated learning to tackle the challenge of non-IID data distribution across clients. 
Yet most federated learning frameworks are based on convolutional neural networks (CNNs), which generally focus on high-frequency local patterns that might be particularly sensitive to data heterogeneity \cite{geirhos2018imagenet}.
Transformers \cite{vaswani2017attention}, however, use self-attention to learn the global interactions between inputs \cite{ramachandran2019stand} and, therefore, they tend to be more robust to distribution shifts and data heterogeneity \cite{bhojanapalli2021understanding}. Motivated by this, a very recent work \cite{qu2022rethinking} proposed the idea of using Transformers as a federated network architecture coupled with the basic federated averaging (FedAvg) algorithm \cite{mcmahan2017communication}. Although the study showed some very promising experimental results, the impacts that federated learning algorithms may have on self-attention has not yet been studied.  It is our fear that such algorithms may limit the capabilities of Transformers in federated learning. Given the promise of Transformer-based federated learning, this is a topic worthy of further investigation. 

To sum up, the main issues of current personalized federated learning are as follows:
\begin{itemize}
\item [1)] Many existing CNNs-based federated models might be particularly sensitive to data heterogeneity since CNNs generally focus on high-frequency local patterns.
\item [2)] Existing Transformer-based Federated learning methods do not investigate the effects of the federated aggregation operations on the self-attention. 
\item [3)] There is a lack of a personalized federated learning framework that is more appropriate for Transformer structures. 
\end{itemize}

Recently, it has been demonstrated that the self-attention layers in Transformers play a more important role than the other layers \cite{bhojanapalli2021understanding}. Inspired by this, we devised a simple experiment to explore the contribution of self-attention to federated learning and to study the effects of federated learning methods on self-attention mechanisms. In our experiment, we compared the attention maps of three different Vision Transformer (ViT) models \cite{dosovitskiy2020image}, these being: 1) Local-T, which trains a unique ViT in each client locally; 2) FedAvg-T, which applies the FedAvg algorithm to train a global ViT; and 3) Vanilla Personalized-T, which retains the self-attention locally and uses FedAvg to aggregate the other parameters in the server. The experiment itself involved sampling a set of images from different classes in CIFAR-100 \cite{krizhevsky2009learning} with five clients and using the Attention Rollout method \cite{abnar2020quantifying} to produce attention maps. The original images and corresponding attention maps of these methods are shown in Fig. \ref{attention_map}. We can see that both Local-T and Vanilla Personalized-T can discover critical information in the images (the red area in the images), but FedAvg-T failed to generate a meaningful attention map. This indicates that aggregating the self-attention of clients with heterogeneous data may ruin the client-specific representations, potentially degrading the model’s performance. 

Unlike applying simple FedAvg operations on the whole model, the above vanilla personalization mechanism can depict client-specific self-attention layers by local training. However, since these personalized self-attention layers are learned independently without considering the potential inherent relationships between clients, the obtained personalized self-attention may be sub-optimal. Moreover, the personalized self-attention layers are not scalable as they increase linearly with the increase of client numbers. Furthermore, the generalization of personalized self-attention to novel clients is limited since the whole self-attention layers must be re-trained. 
Based on this, we designed a novel Transformer-based federated learning framework called Federated Transformer Personalization (FedTP) that uses a 
\emph{learn-to-personalize} mechanism instead of the above vanilla personalization. Specifically, a hypernetwork is learned on the server that generates projection matrices in self-attention layers to produce client-wise queries, keys, and values, while the other model parameters are aggregated and shared.
The \emph{learn-to-personalize} mechanism for self-attention layers through hypernetwork allows us to effectively share parameters across clients and generate personalized self-attention layers by learning a unique embedding vector for each client. In addition to high accuracy, FedTP is also scalable to the increase of client number and enjoys good generalization capability to novel clients. Our main contributions are summarized as follows: 
\begin{itemize}
    \item[1)] We explore the effects of self-attention mechanism in personalized federated learning and we are the first to reveal that FedAvg may have negative impacts on self-attention where data heterogeneity is present. Based on this, we propose a novel Transformer-based federated learning framework, namely FedTP, that learns personalized self-attention for each client.
    \item[2)] We propose a \emph{learn-to-personalize} mechanism to better exploit clients' cooperations in the personalized layers and improve the scalability and generalization of FedTP.  In addition, we derive the generalization bounds for FedTP with the \emph{learn-to-personalize} mechanism.
    \item[3)] We conducted extensive experiments on three benchmark datasets under different non-IID data settings. The experimental results demonstrate that FedTP yields state-of-the-art performance over a wide range of personalized federated learning benchmark methods on both image and language tasks. 
\end{itemize} 

The rest of this paper is organized as follows. Section \uppercase\expandafter{\romannumeral2} briefly reviews the related work of personalized federated learning, Transformers and Hypernetworks. Section \uppercase\expandafter{\romannumeral3} presents the formulation of Transformer-based personalized federated learning; the details of two types of personalization, including vanilla personalization and \emph{learn-to-personalize}; and the update of model parameters. Experimental results and discussions are provided in Section \uppercase\expandafter{\romannumeral4}. Section \uppercase\expandafter{\romannumeral4} analyzes the limitations of FedTP and section \uppercase\expandafter{\romannumeral5} concludes the whole paper. The proof of Theorem 1 in Section \uppercase\expandafter{\romannumeral3} is given in the Appendix. 

\section{Related Work}
\subsection{Personalized Federated Learning}
Multiple approaches to personalized federated learning have been proposed as a way of overcoming heterogeneity across clients \cite{zhang2022r, zhang2022semi,zhang2022federated}. Currently, these methods can be divided into two main categories: global model personalization and learning personalized models \cite{tan2022towards}. Global model personalization aims to improve the performance of a single shared global model given heterogeneous data. An intuitive method is fine-tuning the global model on the clients' local datasets to produce personalized parameters \cite{wang2019federated,mansour2020three,fallah2020personalized,khodak2019adaptive}. Another strategy is to add a proximal regularization term to handle client drift problems resulting from statistical heterogeneity. Examples include FedProx \cite{li2020federated}, pFedMe \cite{t2020personalized} and Ditto \cite{li2021ditto}. Instead, FedAlign \cite{mendieta2022local} studied the data heterogeneity challenge of FL from the perspective of local learning generality rather than proximal restriction.

Instead of training a single global model, learning personalized models is more suitable for heterogeneous clients. One such approach is to seek an explicit trade-off between the global model and the local models. Some researchers have considered interpolating the two models, e.g., L2GD \cite{hanzely2020federated} and LG-FEDAVG \cite{liang2020think}. Similarly, Knn-Per \cite{marfoq2022personalized} is also an interpolation of the two models, but the local model is built through a k-nearest neighbors method, which requires storing all the features of the samples. To increase the flexibility of this personalized model architecture for clients, some methods distill knowledge from a global teacher model into student models on the client devices. In this way, the clients learn a stronger personalized model. FedMD \cite{li2019fedmd} and FedGen \cite{zhu2021data} are two examples of this approach. When there are inherent partitions or differences in the data distribution between clients, it can be more appropriate to train a federated learning model for each homogeneous group of clients through a clustering approach, like CFL \cite{sattler2020clustered}, PFA \cite{liu2021pfa}, or FedAMP \cite{huang2021personalized}, noting, though, that this may lead to high computation and communication overheads. FedPer \cite{arivazhagan2019federated} and FedRep \cite{collins2021exploiting} learn personalized classifier heads locally while sharing the base layers. FedBN \cite{li2020fedbn} updates the client batch normalization layers locally, and pFedGP \cite{achituve2021personalized} learns a shared kernel function for all clients and a personalized Gaussian process classifier for each client. In addition, some studies focus on specific fields and propose corresponding PFL methods. FEDGS \cite{li2022data} is proposed to solve the heterogeneity caused by the rapidly changing streaming data and clustering nature of factory devices in the Industrial Internet of Things. FedStack \cite{shaik2022fedstack} supports ensembling heterogeneous architectural client models in the area of healthcare. Furthermore, a horizontal federated reinforcement learning (HFRL)-based method \cite{zhang2022federated2} develops optimal electric vehicles charging/discharging control strategies for different EVs users to maximise their benefits. Unlike these works, our FedTP employs a \emph{learn-to-personalize} mechanism to learn personalized self-attention layer in Transformer to better handle data heterogeneity among clients. 

\subsection{Transformers}
The Transformer model \cite{vaswani2017attention} was originally designed to improve the efficiency of machine translation tasks. It is a deep learning model basefeaturesself-attention mechanism that has achieved state-of-the-art performance in many NLP tasks. Since it first emerged, many researchers have explored its applicability to vision tasks, with one of the most successful attempts being ViT \cite{dosovitskiy2020image}. As the first attempt to use Transformer in federated learning, Park et al. proposed a federated ViT for COVID-19 chest X-ray diagnosis \cite{park2021federated}. This framework leverages robust representations from multiple related tasks to improve the performance of individual tasks. Very recently, Qu et al. conducted rigorous empirical experiments and showed that, in federated learning settings, Transformers are more suitable for situations with heterogeneous data distributions than CNNs \cite{qu2022rethinking}. However, although Transformer models have shown very promising performance in federated learning, most are only trained with the basic FedAvg algorithm, which may limit their true capabilities, especially in non-IID scenarios. To see this architecture perform to its fullest potential, our FedTP learns a personalized Transformer for each client. The self-attention mechanism captures data heterogeneity through a hypernetwork located on the server, which generates projection parameters in the self-attention layers. 

\subsection{Hypernetworks}
Hypernetworks \cite{ha2016hypernetworks} are neural networks that can generate weights for a large target network with a learnable embedding vector as its input. pFedHN \cite{shamsian2021personalized} was the first method to apply a hypernetwork to personalized federated learning, where a hypernetwork is learned on the server that generates personalized weights for the local CNNs on each client. Instead of generating all the model’s parameters, pFedLA \cite{ma2022layer} employs a hypernetwork on the server that outputs aggregated weights for each layer of the local model on each different client. Different from these two studies, Fed-RoD \cite{chen2022bridging} learns a hypernetwork locally, which outputs personalized predictors for clients with the extra inputs of the clients' class distributions. Notably, all these methods with hypernetworks are based on CNN architectures. In contrast, the hypernetwork in FedTP generates projection matrices in the self-attention layers of a Transformer to produce client-wised queries, keys, and values. 

\section{Federated learning by Transformer Personalization}
In this section, we present the design of our FedTP framework. Aiming to mitigate heterogeneity and build a high-quality personalized model for each client, FedTP learns personalized self-attention layers for each client.

\begin{figure*}[t]
\centering
\includegraphics[width=0.95\textwidth]{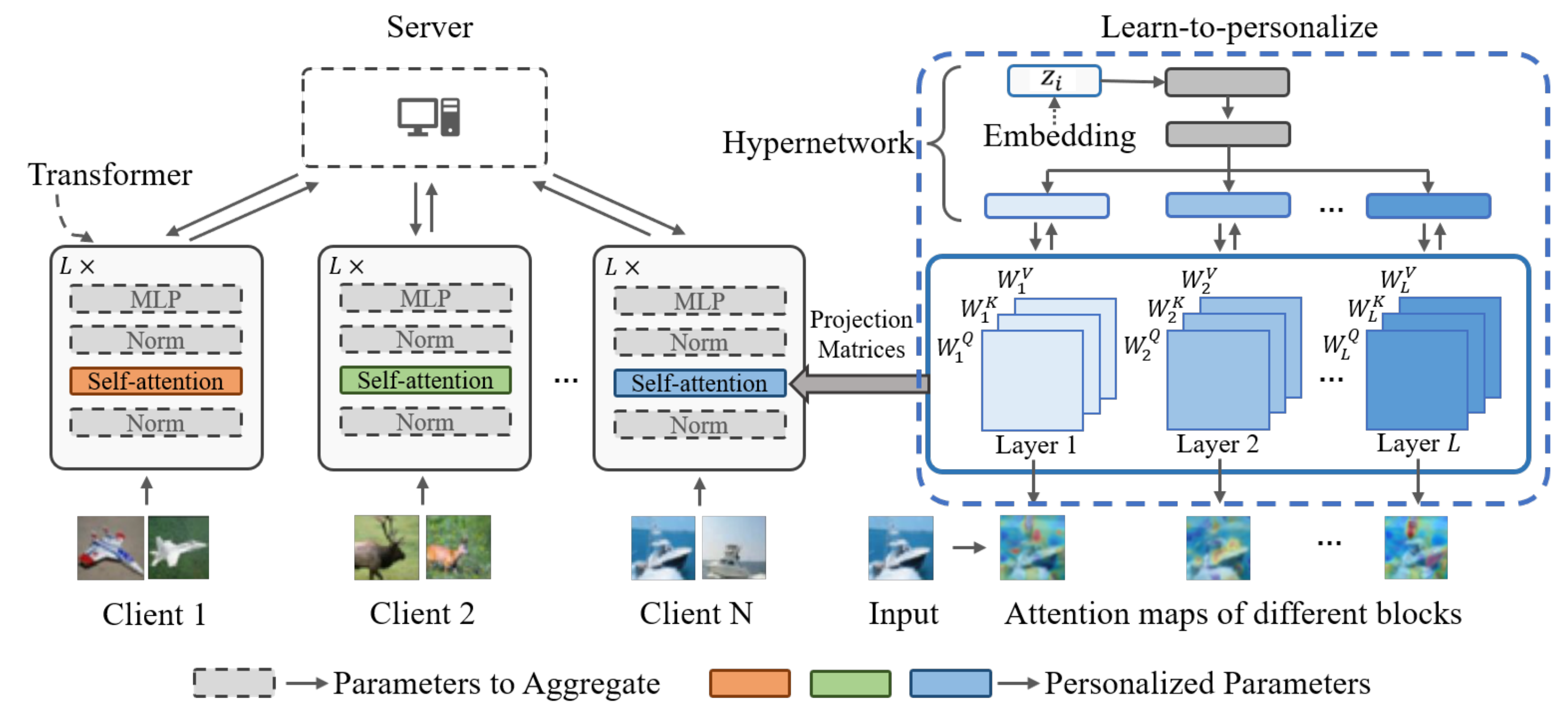} 
\caption{An overview of FedTP. On the left, the self-attention layers are retained locally while the other parameters are aggregated on the server and shared among clients. 
The right shows the workflow of {\it learn-to-personalize} that generates the projection matrices in the self-attention layers of $L$ Transformer blocks. This is performed by a hypernetwork located on the server, which consists of simple MLP layers with the last layer being unique for each Transformer block.}
\label{pipline}
\end{figure*}

\subsection{Problem Formulation}
The FedTP framework uses a traditional ViT \cite{dosovitskiy2020image} for image tasks, and Transformer \cite{vaswani2017attention} for language tasks. The input sequence $X$ is with a predetermined length of $m$, noting that images are partitioned into a sequence in the image pre-pocessing layer of ViT. This sequence is then transformed into a corresponding embedding matrix $H$ with a dimension of $\mathbb{R}^{m \times d}$. The queries, keys and values of the self-attention mechanism are respectively denoted as $Q=HW^Q$, $K=HW^K$ and $V=HW^V$. For convenience, we concatenated these projection parameters as $W=[W^Q,W^K,W^V]$. Next, self-attention is applied through $Attention(Q,K,V)=softmax(\frac{QK^T}{\sqrt{d}})V$, where $d$ is the number of columns of $Q$, $K$ and $V$. 

To simulate a federated scenario, we assume there are $N$ clients indexed by $[N]$ and each client $i$ has a local dataset ${D}_i=\{(x_i^{(j)},y_i^{(j)})\}_{j=1}^{m_i} (1 \leq i \leq N)$ with $m_i$ samples drawn from a distinct data distribution $\mathcal{P}_i$. Let ${D}=\bigcup_{i\in [N]} {D}_i$ denote the total datasets with the size of $M=\sum_{i=1}^N m_i $, and let $f(\theta_i;\cdot)$ denote a personalized model for client $i$, parameterized by $\theta_i$. The optimization objective is: 
\begin{equation}\label{eq1}
    \centering
    \mathop{\arg\min}\limits_{\Theta} \sum_{i=1}^{N} \frac{m_i}{M}\mathcal{L}_i(\theta_i),
\end{equation}
with 
$\mathcal{L}_i(\theta_i) = \mathbb{E}_{(x_i^{(j)},y_i^{(j)})\in {D}_i}l(f(\theta_i;x_i^{(j)}),y_i^{(j)})$, 
where $\Theta=\{\theta_i\}_{i=1}^N$ is the set of personalized parameters, and $l(\cdot,\cdot)$ denotes the per sample loss function common to all clients. The loss function can be Mean Square Error or Cross Entropy Loss, which is selected according to different tasks.

\subsection{Vanilla personalization for self-attention} 
Transformer-based federated learning has received increasing attention. It is well known that Transformer can capture global interactions between the inputs by the self-attention layers.  As we analyzed in Introduction, simply applying the FedAvg operation on the self-attention layers of clients may degrade the model's performance in data heterogeneity scenarios. Inspired by this, we first develop a personalized self-attention mechanism that maintains the self-attention layers of each client locally so as to personalize the model, while averaging other parameters to learn the common information. The left side of Fig. \ref{pipline} shows the structure of vanilla personalization for self-attention. 

Similar to FedAvg, these parameters are updated via local training for several epochs and aggregated on the server. Let $W_i$ denote the projection parameters of the self-attention layer, and let $\xi$ denote the parameters of other layers. The personalized parameter $\theta_i$ is then be split into $\theta_i = \{W_i,\xi \}$. During the communication round $t$, the personalized model $f(\theta_i;\cdot)=f(W_i,\xi;\cdot)$ is trained locally for $k$ steps, and the model becomes $f(W_i^{t,k}, \bar{\xi}_i^{t,k};\cdot)$, where $W_i^{t,k}$ is retained locally to store the personalized information of client $i$ and $\bar{\xi}_i^{t,k}$ is aggregated across the clients via
\begin{equation}\label{eq3}
    \centering
\bar{\xi}^{t}=\sum_{i=1}^{N} \frac{m_i}{M}\xi_i^{t,k}.
\end{equation}
Thus, the objective function of FedTP can be derived from Eq. (\ref{eq1}) to minimize the following loss:
\begin{equation}\label{eq4}
    \centering
    \mathop{\arg\min}\limits_{\Theta} \sum_{i=1}^{N} \frac{m_i}{M} \mathcal{L}_i (W_i,\xi),
\end{equation}
where 
\begin{equation}
    \mathcal{L}_i (W_i,\xi) = \sum\limits_{i=1}^{N} \dfrac{m_i}{M} \mathbb{E}_{(x_i^{(j)},y_i^{(j)})\in {D}_i}l(f(W_i,\xi; x_i^{(j)}),\\ y_i^{(j)}).
\end{equation} 

The above vanilla personalization procedure can generate personalized self-attention for each client by local training. However, since it ignores the potential inherent relationships between clients, the obtained personalized self-attention may be stuck in sub-optimality. In addition, the personalized self-attention layers are not scalable as they increase linearly with the growing client numbers. Moreover, the generalization of personalized self-attention is not good. For example, when novel clients are involved, the model needs re-training to generate the specific self-attention layers for these novel clients. 

\subsection{Learn-to-personalize for self-attention}
{\color{black} In this section, we develop FedTP that utilizes the \emph{learn-to-personalize} mechanism to improve the vanilla personalization mechanism for self-attention.} FedTP learns a  hypernetwork\cite{ha2016hypernetworks} at the server to generate projection matrices in the self-attention layers for each client (see the right side of Fig. \ref{pipline}). By this way, FedTP can effectively share parameters across clients and maintain the flexibility of personalized Transformers. 

The hypernetwork on the server is denoted as $h(\varphi;z_i)$ parameterized by $\varphi$, where $z_i \in \mathbb{R}^D$ can be a learnable embedding vector corresponding to client $i$ or fixed. We implement the hypernetwork with simple fully-connected layers and the last layer is unique for each Transformer block. Given the embedding vector $z_i$, the hypernetwork outputs the partial weights $W_i=h(\varphi;z_i)$ for client $i$, which is then decomposed into the projection parameters for the queries, keys and values of the self-attention mechanism $W_i=[W_i^Q, W_i^K, W_i^V]$. By this way, the hypernetwork learns a category of projection parameters $\{W_i = h(\varphi;z_i) | 1 \leq i \leq N  \}$ for personalized self-attention. Hence, the personalized model is denoted as $f(W_i,\xi;\cdot)=f(h(z_i;\varphi),\xi;\cdot)$ and the training loss in Eq. (\ref{eq4}) is replaced by
\begin{equation}\label{eq5}
    \centering
    \begin{aligned}
    & \mathcal{L}_i (W_i,\xi) = \mathcal{L}_i (h(\varphi;z_i),\xi) \\
    &=  \sum_{i=1}^{N} \frac{m_i}{M} \mathbb{E}_{(x_i^{(j)},y_i^{(j)})\in {D}_i}l(f(h(\varphi;z_i),\xi;x_i^{(j)}),y_i^{(j)}).
    \end{aligned}
\end{equation}

\begin{algorithm}[t]
\label{alg:algorithm}
\caption{Federated Transformer Personalization}
\LinesNumbered
        \KwIn{$T-$ number of communication rounds, $K-$ number of local epochs, $\alpha-$ learning rate of local update, $\beta-$ learning rate for global update. }
        Initialize parameters $\xi^0$, $z_i^0$ and $\varphi^0$; \\
        \For{each communication rounds $t \in \{1,...,T\}$}
        {
            Sample the set of clients $C^t \subset \{1,...,N\}$; \\
            \For{each client $i \in C^t$}
            {
                $\xi^{t,0}=\bar{\xi}^{t-1}$; \\
                $W_i^{t,0}=h(\varphi^{t-1};z_i^{t-1})$; \\
                $\theta_i^{t,0}=\{W_i^{t,0},\xi^{t,0}\}$; \\
                    \For{each local epoch $k \in \{1,...,K\}$}
                    {Sample mini-batch $B_i \in D_i$; \\
                    $\theta_i^{t,k+1} \leftarrow \theta_i^{t,k} - \alpha \nabla_{\theta_i}\mathcal{L}_i(\theta_i^{t,k};B_i)$; 
                    }
                $\Delta W_i = W_i^{t,K}-W_i^{t,0}$; 
                    }
            $\bar{\xi}^t = \sum_{i \in C^t} \frac{m_i}{M}\xi^{t,K}$; \\
            $\varphi^t = \varphi^{t-1} - \beta \sum_{i \in C^t} \frac{m_i}{M} \nabla_{\varphi}W_i^T \Delta W_i$; \\ 
            $z_i^t = z_i^{t-1}-\beta \sum_{i \in C^t} \frac{m_i}{M} \nabla_{z_i}W_i^T \Delta W_i$; 
            }
         \textbf{return} $\bar{\xi}^t$, $\varphi^t$ and $z_i^t$    
\end{algorithm}

Algorithm \ref{alg:algorithm} demonstrates the procedures of FedTP algorithm. We next introduce the update rules for model parameters in FedTP. First, in each local epoch $k$ we update local model parameter $\theta_i$ using stochastic gradient descent (SGD) by
\begin{equation}\label{equ6}
    \theta_i^{k} \leftarrow \theta_i^{k-1} - \alpha   \nabla_{\theta_i}\mathcal{L}_i(\theta_i^{k-1};B_i),
\end{equation}
where $B_i$ is a mini-batch sampled from $D_i$. Let $C^t$ represent the set of sampled clients at each round $t$. According to the chain rule, we can obtain the gradients of $\varphi$ and $z_i$ from Eq. (\ref{equ6}):
\begin{equation}\label{equ7}
    \nabla_\varphi \mathcal{L}_i =\sum_{i \in C^t} \frac{m_i}{M} \nabla_{\varphi}W_i^T \Delta W_i,
\end{equation}
\begin{equation}\label{equ8}
    \nabla_{z_i} \mathcal{L}_i =   \sum_{i \in C^t} \frac{m_i}{M} \nabla_{z_i}W_i^T \Delta W_i,
\end{equation}
where $\Delta W_i = W_i^K - W_i^0 $ is the change of projection parameters after $K$ epochs, and $K$ is the local updating epochs. In communication round $t$, we update hypernetwork parameter $\varphi$ and client embedding $z_i$ by using the calculated gradients:
\begin{equation}\label{equ10}
\varphi^{t} = \varphi^{t-1} - \beta \nabla_\varphi \mathcal{L}_i^{t-1},
\end{equation}
\begin{equation}\label{equ11}
 z_i^{t} = z_i^{t-1} - \beta \nabla_{z_i} \mathcal{L}_i^{t-1}.  
\end{equation}

Compared to the vanilla personalization mechanism, \emph{learn-to-personalize} for self-attention has several merits: 1) it can effectively share parameters across clients and better exploits the role of self-attention mechanism in federated learning; 2) it offers scalability with the growing number of clients since the self-attention layer is generated by the shared hypernetwork with client-wise embedding vectors; 3) it can be better generalized to novel clients whose data is unseen during training. The first and third points will be validated in Section IV. The 
second point can be easily shown by comparing the total number of parameters of \emph{learn-to-personalize} and vanilla personalization. Based on the MLP structure of hypernetwork, its total number of parameters is approximately equal to $D_h \times D_s$, where $D_h$ and $D_s$ are the dimensions of the hidden layers in hypernetwork and the projection parameters of self-attention. In contrast, the total projection parameters of vanilla personalization increase linearly with the number of clients, i.e. $N \times D_s$. When the client number $N$ is larger than $D_h$, {\it learn-to-personalize} for self-attention consumes less resource consumption.

\subsection{Generalization Bound}
We analyze the generalization bound of FedTP in this section. Before starting the analysis, we first introduce some assumptions as follows. 

\noindent \textbf{Assmuption 1}
\emph{We assume the embedding vectors $z_i$ and the weights $\varphi$ of hypernetwork $h(\varphi,z_i)$ are bounded in a ball of radius $R_h$, and the parameters of other layers $\xi$ in Transformer are bounded in a ball of radius $R_t$. These assumptions can be denoted as:
\begin{equation}
\Vert \varphi-\varphi^{\prime} \Vert \leq R_h,\ \Vert z_i-z_i^{\prime} \Vert \leq R_h,\ \Vert \xi-\xi^{\prime} \Vert \leq R_t.
\end{equation}
}

\noindent \textbf{Assmuption 2}
\emph{(Lipschitz conditions) Let $\mathcal{D}_1,\mathcal{D}_2,\cdots,\mathcal{D}_N$ represent the real data distributions, and $\mathcal{L}_{\mathcal{D}_i}(h(\varphi;z_i ),\xi)=\mathbb{E}_{(x,y)\in \mathcal{D}_i}l(f(h(\varphi;z_i),\xi;x),y)$ be the expected loss. We assume the following Lipschitz conditions hold:
}
\begin{subequations}
\begin{equation}
|\mathcal{L}_{\mathcal{D}_i}(h(\varphi;z_i),\xi)-\mathcal{L}_{\mathcal{D}_i}(h(\varphi;z_i),\xi^{\prime})|  \leq L_{\xi} \Vert \xi-\xi^{\prime} \Vert,
\end{equation}
\begin{equation}
\begin{aligned}
&|\mathcal{L}_{\mathcal{D}_i}(h(\varphi;z_i),\xi)-\mathcal{L}_{\mathcal{D}_i}(h^{\prime}(\varphi;z_i),\xi)| \\
&\leq L_h \Vert h(\varphi;z_i)-h^{\prime}(\varphi;z_i)\Vert,
\end{aligned}
\end{equation}
\begin{equation}
\Vert h(\varphi;z_i)-h(\varphi^{\prime};z_i)\Vert \leq L_{\varphi} \Vert \varphi-\varphi^{\prime} \Vert,
\end{equation}
\begin{equation}
\Vert h(\varphi;z_i)-h(\varphi;z_i^{\prime})\Vert \leq L_z \Vert z_i-z_i^{\prime} \Vert.
\end{equation}
\end{subequations}

\noindent \textbf{Theorem 1}
\emph{Suppose ${\mathcal{\hat{D}}}_1, {\mathcal{\hat{D}}}_2,\cdots,{\mathcal{\hat{D}}}_N$ denote the empirical data distribution of $N$ clients with $\hat{\varphi}$, $\hat{z}_i$, $\hat{\xi}$ the parameters learnt by the corresponding empirical distributions. Denote $\mathcal{H}$ as the personalized hypothesis and $d$  be the VC-dimension of $\mathcal{H}$. Suppose Assumptions 1 and 2 hold, with probability at least $1-\delta$, we have  
}
\begin{equation}\label{Generalization Bound}
\centering
\begin{aligned}
    &\left|\sum_{i=1}^{N}\frac{m_i}{M}\mathcal{L}_{\hat{\mathcal{D}}_i}(h(\hat{\varphi};\hat{z_i}),\hat{\xi}) - \sum_{i=1}^{N}\frac{m_i}{M}\mathcal{L}_{\mathcal{D}_i}(h(\varphi^*;z_i^*),\xi^*)\right| \\
    &\leq \sqrt{\frac{M}{2}log\frac{N}{\delta}} + \sqrt{\frac{dN}{M}log\frac{eM}{d}} +L_h R_h (L_{\varphi}+L_z) \\
    &+L_{\xi} R_t,
\end{aligned}
\end{equation}
\noindent\emph{where 
$\varphi^*$, $z_i^*$, and $\xi^*$ represent the optimal parameters corresponding to the real distribution of each client, respectively; the size of the whole dataset is $M$ with the local data size of client $i$ being $m_i$. 
}

Theorem 1 indicates that the performance of the model trained on the empirical distribution is affected by the complexity of the hypothesis class $\mathcal{H}$ (expressed by its VC-dimension), the number of clients, the size of the whole datasets, and the Lipschitz constants. 
{\color{black} The second part on the right-hand side of (\ref{Generalization Bound}) can be formulated as $\sqrt{\frac{log (eM/d)}{M/dN}}$. It means that it is closely related to $\frac{M}{d}$. We denote the hypothesis class of FedTP with \emph{learn-to-personalize} and with vanilla personalization by $\mathcal{H}_h$ and $\mathcal{H
}_v$, respectively. The VC-dimension of $\mathcal{H}_h$ is smaller than the VC-dimension of $\mathcal{H}_v$ especially for large number of clients since we use one hypernetwork to generate the self-attention layers for all clients with the \emph{learn-to-personalize} mechanism. With the reduction of the VC-dimension $d$, $\sqrt{\frac{log (eM/d)}{M/dN}}$ will decrease. Thus, FedTP owns better generalization than vanilla personalization for self-attention especially when there are a large number of clients.}
The key lemmas and proof of Theorem 1 are given in the Appendix.

\section{Experiments}
\begin{figure}[h]
\centering
\includegraphics[width=0.72\columnwidth]{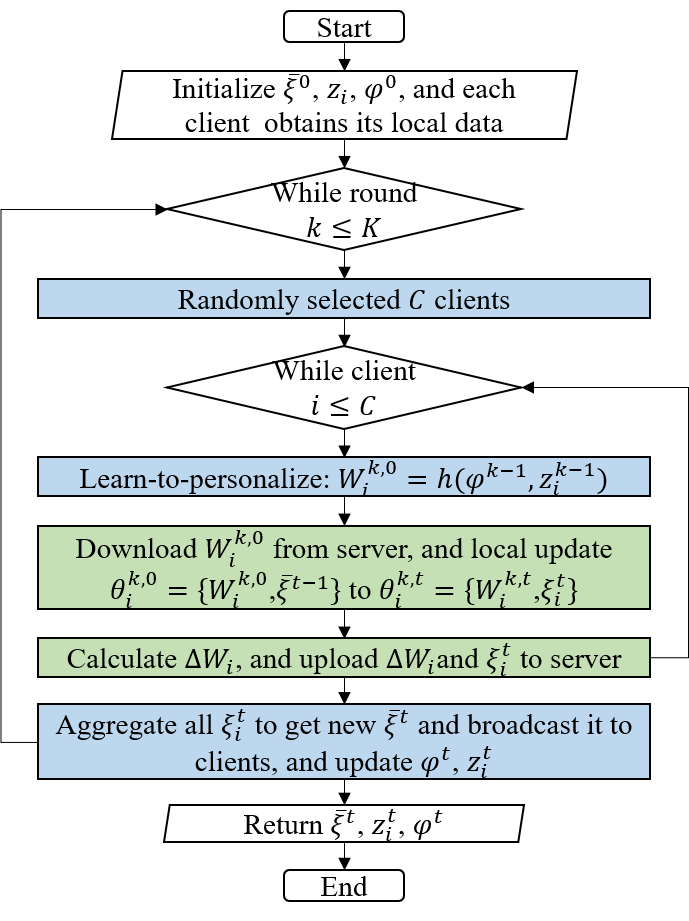}
\caption{The flow chat of FedTP. The blue part of the flow chart represents the operations that occur on the server side, while the green part indicates the client-side operations.}
\label{flow_chart}
\end{figure}

In this section, we describe the setup of experiments, evaluate the performance of our model and compare it to several baseline methods in various learning setups. In the Experimental Setup Subsection, we introduce benchmarks we compared, Non-IID settings and model architectures we apply, and some implementation details. In the Performance Evaluation part, we mainly analyze the performance of different network backbones and different personalized parts of Transformer, generalization to novel clients, compatibility of FedTP to other methods, and visualization of attention maps. In Ablation Study Subsection, we study the effect of heterogeneity in label distribution and in Noise-based Feature Imbalance, and the impact of different parameters. Fig. \ref{flow_chart} shows the flow chart of our FedTP.

\begin{table*}[t]
\renewcommand\arraystretch{1.25}
\centering
\caption{Datasets and models.}
\label{Datasets_and_models}
\begin{tabular}{lllrl}
\hline
\multicolumn{1}{c}{Dataset} & \multicolumn{1}{c}{Task}  & \multicolumn{1}{c}{Clients} & \multicolumn{1}{c}{Total Samples} & \multicolumn{1}{c}{Model} \\ \hline
CIFAR-10                     & Image Classification      & 50/100                   & 60,000                            & ConvNet, ViT        \\
CIFAR-100                    & Image Classification      & 50/100                   & 60,000                            & ConvNet, ViT        \\
Shakespeare                 & Next-character Prediction & 683                         & 2,578,349                         & LSTM, Transformer               \\ \hline
\end{tabular}
\end{table*}

\subsection{Experimental Setup}

\emph{1) Benchmarks:} We compared FedTP with the basic Federated algorithms, such as \textbf{FedAvg} \cite{mcmahan2017communication} and \textbf{FedProx} \cite{li2020federated}, as well as with several advanced personalization algorithms: \textbf{FedPer} \cite{arivazhagan2019federated}, \textbf{pFedMe} \cite{t2020personalized}, \textbf{FedBN} \cite{li2020fedbn}, \textbf{pFedHN} \cite{shamsian2021personalized}, \textbf{pFedGP} \cite{achituve2021personalized}, 
and \textbf{FedRoD} \cite{chen2022bridging}. 

\emph{2) Non-IID Settings of Datasets:} 
We use three popular benchmark datasets: CIFAR-10, CIFAR-100 \cite{krizhevsky2009learning}, and Shakespeare\cite{caldas2018leaf, mcmahan2017communication}. The first two are image datasets and the last one is a language dataset.
For CIFAR-10 and CIFAR-100, we applied two split strategies to simulate non-IID scenarios. The first is a ``Pathological'' setting, where each client is randomly assigned two/ten classes for CIFAR-10/CIFAR-100 as in \cite{t2020personalized}. The sample rate on client $i$ of selected class $c$ is obtained by ${a_{i,c}}/{\sum_{j} a_{j,c}}$, where $a_{i,c} \sim U(.4,.6)$.
The second is a federated version created by randomly partitioning the datasets among clients using a symmetric Dirichlet distribution with parameter $\alpha=0.3$, as in \cite{reddi2020adaptive, marfoq2022personalized}. We create federated versions of CIFAR-10 by randomly partitioning samples with the same label among clients according to a Dirichlet distribution with parameter $\alpha=0.3$. As for CIFAR-100, in order to create more realistic local datasets, we use a two-stage Pachinko allocation method to partition samples over ``coarse'' and ``fine'' labels. This method firstly generates a Dirichlet distribution with parameter $\alpha=0.3$ over the coarse labels for each client, and then generates a Dirichlet distribution with parameter $\beta=10$ over the coarse corresponding fine labels. For both partitions, the classes and distribution of classes in training and test set of each client are the same.
For Shakespeare, similar to the partitions in \cite{marfoq2022personalized}, we maintained the original split between the training and testing data specifically, 80$\%$ for training and 20$\%$ for testing. 
Table \ref{Datasets_and_models} summarizes the datasets, corresponding tasks, and the number of clients and models.

\emph{3) Model Architectures:} 
Similar to many federated works \cite{shamsian2021personalized,chen2022bridging}, we adopted a ConvNet \cite{ConvNet} with 2 convolutional layers and 3 fully-connected layers as the local neural networks for baseline methods on CIFAR-10 and CIFAR-100. To improve the communication efficiency of large-scale federated scenarios, we chose a tiny ViT with fewer parameters for FedTP, which consisted of 8 blocks with 8 self-attention layers in each block. The corresponding attention head number is 8, and the patch size is 4 and the hidden size is 128.
With regards to Shakespeare, we applied the same stacked two-layer LSTM model as \cite{marfoq2021federated,marfoq2022personalized} for benchmark methods. 
For a fair comparison with LSTM, we used a simple Transformer \cite{vaswani2017attention} with two blocks as an encoder. The depth for both the LSTM and the simple Transformer is 2 and the inner hidden dimension for both is 256. 
Since FedTP has a similar backbone for image classification tasks and language prediction tasks, we are able to use the same structure for the hypernetwork $h(\varphi;\cdot)$ except for the last layer.
We implement this hypernetwork similar to \cite{shamsian2021personalized} with an MLP network and parameter mapping heads. The MLP network consists of four fully-connected layers with 150 neurons and each parameter mapping head is a single FC layer.
For FedBN \cite{li2020fedbn}, we follow its original design which adds one BN layer after each convolution layer and fully-connected layer (except for the last layer). For FedRoD \cite{chen2022bridging}, since the reported accuracy between hypernetwork version and the linear version is similar, we adopt the linear version.

\begin{table*}[t]
\renewcommand\arraystretch{1.25}
\caption{Average test accuracy for FedTP and several Transformer-based methods with different non-iid settings. 
}
\label{transformer-based}
\resizebox{1\textwidth}{!}{
\centering
\begin{tabular}{lllllllll}
\hline
            & \multicolumn{4}{c}{CIFAR-10}                                                                                                                & \multicolumn{4}{c}{CIFAR-100}                                                                                                               \\ \cline{2-9} 
\#setting   & \multicolumn{2}{c}{Pathological}                                     & \multicolumn{2}{c}{Dirichlet}                                        & \multicolumn{2}{c}{Pathological}                                     & \multicolumn{2}{c}{Dirichlet}                                        \\ \cline{2-9} 
\#client    & \multicolumn{1}{c}{50}                     & \multicolumn{1}{c}{100} & \multicolumn{1}{c}{50}                     & \multicolumn{1}{c}{100} & \multicolumn{1}{c}{50}                     & \multicolumn{1}{c}{100} & \multicolumn{1}{c}{50}                     & \multicolumn{1}{c}{100} \\ \hline
Local-T     & 84.55±0.15         & 82.21±0.08              & 69.94±0.13          & 66.68±0.13              & 55.91±0.17          & 49.25±0.11              & 27.87±0.12          & 23.34±0.10              \\
FedAvg-T \cite{mcmahan2017communication}   & 50.42±4.22          & 46.28±4.23              & 61.85±1.5           & 59.23±1.93              & 34.02±0.88          & 30.20±0.95              & 38.64±0.22          & 34.89±0.45              \\
FedProx-T \cite{li2020federated}  & 47.28±5.09          & 44.22±3.81              & 61.00±1.74          & 60.13±1.71              & 29.77±1.67          & 28.92±0.83              & 37.63±0.35          & 32.98±0.43              \\
FedPer-T \cite{arivazhagan2019federated}   & 89.86±0.89         & \textbf{89.01±0.12}     & 79.41±0.16         & 77.70±0.14              & 67.23±0.32          & 61.72±0.16              & 37.19±0.18          & 29.58±0.14              \\ 
pFedMe-T \cite{t2020personalized}   & 82.26±0.61          & 77.57±0.52              & 71.45±0.87          & 68.13±0.67              & 53.08±0.72          & 39.94±0.91              & 33.59±0.52          & 25.95±0.64              \\ \hline
Vanilla Personalized-T    & 86.97±0.07          & 84.90±0.11     & 75.14±0.12          & 72.33±0.16              & 61.98±0.13         & 57.60±0.12              & 37.21±0.11          & 32.81±0.11              \\  
FedTP & \textbf{90.31±0.26} & 88.39±0.14              &  \textbf{81.24±0.17} & \textbf{80.27±0.28}     & \textbf{68.05±0.24} & \textbf{63.76±0.39}     & \textbf{46.35±0.29} & \textbf{43.74±0.39}     \\ \hline
\end{tabular}
}
\end{table*}

\emph{4) Implementation Details:} 
Following the experimental setting in pFedHN \cite{shamsian2021personalized}, we implement FedTP and the benchmark methods with 50 and 100 clients at 10$\%$ and 5$\%$ participation for CIFAR-10 and CIFAR-100, respectively. In the Shakespeare scenario, we treated each identity as a client and sampled 10$\%$ clients in each communication round. 
For the image classification task, we trained every algorithm for 1500 communication rounds. To make for an equivalent communication cost, pFedHN \cite{shamsian2021personalized} is trained for 5000 global communication rounds.
For the next-character prediction task, we trained the corresponding methods for 300 communication rounds. Both tasks were optimized with a SGD optimizer for 5 local epochs with a default learning rate of $lr = 0.01$ and a batch size of $B=64$.
In FedTP and pFedHN, we optimize the hypernetworks by using the SGD optimizer with a default learning rate $\beta$ = 0.01. 
The server and all clients are simulated on a cluster with an RTX 2080 Ti GPU, and all algorithms are implemented in PyTorch \cite{2019pytorch}. 

\subsection{Performance Evaluation}
\emph{1) Evaluation Protocol:} 
We tested each algorithm every 5 rounds during its last 200 global communication rounds and computed the mean test accuracy and standard deviation over these evaluation steps to estimate the model performance. The average test accuracy in each evaluation step is defined as the ratio of the sum of each client's positive predictions over all testing image numbers.

\begin{table*}[thbp]
\centering
\caption{The results of FedTP and the benchmark methods on the image datasets with different non-IID settings.}
\label{results}
\renewcommand\arraystretch{1.25}
\resizebox{1\textwidth}{!}{
\begin{tabular}{lcccccccc}
\hline
            & \multicolumn{4}{c}{CIFAR-10}                                                               & \multicolumn{4}{c}{CIFAR-100}                                                               \\ \cline{2-9}
\#setting   & \multicolumn{2}{c}{Pathological}                 & \multicolumn{2}{c}{Dirichlet}               & \multicolumn{2}{c}{Pathological}                 & \multicolumn{2}{c}{Dirichlet}                \\ \cline{2-9}
\#client    & 50                   & 100                  & 50                   & 100                  & 50                   & 100                  & 50                   & 100                   \\ \hline
FedAvg \cite{mcmahan2017communication}      & 47.79±4.48*           & 44.12±3.10*           & 56.59±0.91           & 57.52±1.01           & 15.71±0.35*           & 14.59±0.40*           & 18.16±0.58           & 20.34±1.34            \\
FedProx \cite{li2020federated}     & 50.81±2.94*           & 57.38±1.08*           & 58.51±0.65           & 56.46±0.66           & 19.39±0.63*           & 21.32±0.71*           & 19.18±0.30           & 19.40±1.76             \\
FedPer \cite{arivazhagan2019federated}     & 83.39±0.47*           & 80.99±0.71*           & 77.99±0.02           & 74.21±0.07           & 48.32±1.46*           & 42.08±0.18*           & 22.60±0.59           & 20.06±0.26               \\
pFedMe \cite{t2020personalized}     & 86.09±0.32*           & 85.23±0.58*           & 76.29±0.44           & 74.83±0.28           & 49.09±1.10*           & 45.57±1.02*           & 31.60±0.46           & 25.43±0.52            \\
FedBN \cite{li2020fedbn}      & 87.45±0.95           & 86.71±0.56           & 74.63±0.60           & 75.41±0.37           & 50.01±0.59           & 48.37±0.56           & 28.81±0.50           & 28.70±0.46            \\
pFedHN \cite{shamsian2021personalized}     & 88.38±0.29*           & 87.97±0.70*           & 71.79±0.57           & 68.36±0.86           & 59.48±0.67*           & 53.24±0.31*           & 34.05±0.41           & 29.87±0.69            \\
pFedGP \cite{achituve2021personalized}     & 89.20±0.30*           & 88.80±0.20*           & —                    & —                    & 63.30±0.10*           & 61.30±0.20*           & —                    & —                     \\
FedRoD \cite{chen2022bridging}     & 89.87±0.03           & \textbf{89.05±0.04}  & 75.01±0.09           & 73.99±0.09           & 56.28±0.14           & 54.96±1.30           & 27.45±0.73           & 28.29±1.53             \\ \hline
FedTP & \textbf{90.31±0.26}  & 88.39±0.14           & \textbf{81.24±0.17}  & \textbf{80.27±0.28}  & \textbf{68.05±0.24}  & \textbf{63.76±0.39}  & \textbf{46.35±0.29}  & \textbf{43.74±0.39}     \\ \hline
\end{tabular}
}
\begin{tablenotes}
\item[1] We use * to represent the results reported in previous works \cite{shamsian2021personalized,achituve2021personalized} under the same experimental settings.
\end{tablenotes}
\end{table*}

\emph{2) Performance Analysis:} 
We compared the performance of FedTP to some well-known federated learning methods. Note that these methods were originally implemented based on CNN backbones. Here we used the same Transformer architecture of FedTP to replace their CNN backbones for a fair comparison. To mark the modification, we added a ``-T'' after the algorithm name to denote these methods. The average test accuracy of these algorithms is reported in Table \ref{transformer-based}, which shows that FedTP outperforms all of them by a clear margin. This also validates our prior claims in the Introduction: 1) the FedAvg algorithm may impair the client-specific representations in Transformers as Local-T works even better than FedAvg-T; 2) the personalized self-attention learned by FedTP can handle the data heterogeneity effectively. 
As shown in Table \ref{transformer-based}, FedTP significantly outperforms Vanilla Personalized-T in all settings, which verified that {\it learn-to-personalize} leverages the strengths of self-attention in Transformer models to improve their performance in federated learning over heterogeneous data.
Additionally, FedTP is also superior to FedPer-T. This indicates that personalizing the projection matrices in self-attention layers is much more effective than personalizing the last classification head. 

\begin{figure}[h]
\centering
\includegraphics[width=0.88\columnwidth]{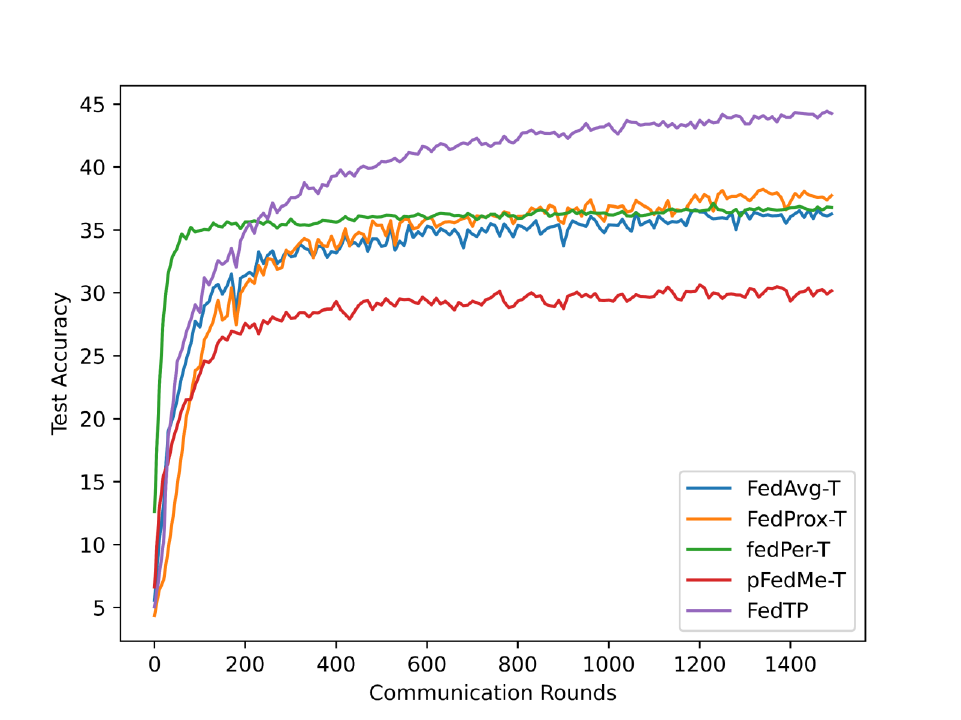} 
\caption{The test accuracy and convergence behavior of FedTP and other Transformer-based Methods on CIFAR-100 with Dirichlet setting over 50 clients.}
\label{dirichlet_CIFAR-100}
\end{figure}

We analyzed the test accuracy vs. global communication rounds curve of FedTP in comparison with other Transformer-based methods in Fig. \ref{dirichlet_CIFAR-100}. FedTP shows a smoother curve and achieves higher accuracy compared with the others. This further proves that the self-attention mechanism is a crucial part to overcome data heterogeneity effectively.
 
\begin{table}[t]
\caption{
Average test accuracy on the language dataset Shakespeare of different methods.
}
\label{shakes_result}
\resizebox{1\columnwidth}{!}{
\centering
\begin{tabular}{lccccc}
\hline
\multicolumn{1}{c}{Method} & FedAvg     & FedProx    & FedPer     & pFedMe     & FedTP      \\ \hline
Test accuracy              & 21.34±1.04 & 20.48±1.09 & 27.56±0.65 & 21.14±1.12 & \textbf{84.40±0.10} \\ \hline
\end{tabular}
}
\end{table}

\emph{3) Analysis of Network Backbone:}
We also implemented experiments to compare the performance of FedTP to many state-of-the-art federated learning methods that are based on CNN/LSTM.
The average test accuracy of these algorithms for the image datasets and language dataset are shown in Table \ref{results} and Table \ref{shakes_result}, respectively. Clearly, FedTP outperforms the baseline methods in terms of the average test accuracy for most cases. It is worth noting that, the hypernetworks in FedTP only produce the parameters of the self-attention layers, which take up around 9.8\% parameters of the whole network. Although FedTP uses a much smaller ratio of personalized parameters, it still significantly outperforms pFedHN with an 11.31\% improvement on average for CIFAR-100. As shown in Table \ref{results} and Table \ref{shakes_result}, changing the network backbone from a CNN or an LSTM to Transformers will bring a significant improvement in the models' ability to overcome data heterogeneity.
Through comparing Table \ref{transformer-based} and Table \ref{results}, the decrease in the accuracy of pFedMe-T and FedProx-T in some settings indicates that the performance of Transformer-based models will be affected when there is a regularization term in these federated learning methods. This is possible because the regularization term may impair the self-attention mechanism, while our FedTP maintains the personalization of the self-attention layers through the {\it learn-to-personalize} mechanism. More details about the effects of {\it learn-to-personalize} will be shown in {\it 5) Visualization of Attention Maps}.

\emph{4) Analysis of Different Personalized Parts:}
Here we investigate the impacts of personalizing different components in the Transformer: (1) the self-attention layers (our method), (2) the MLP layers, (3) the normalization layers, (4) the whole encoder. In this experiment, we use the same hypernetwork to generate the parameters of these different components, and keep the same structures of ViT as we described in the Subsection of Implementation Details. The results are shown in Table \ref{personalized_different_part}, from which we can see that personalizing self-attention layers achieves the best results compared to personalizing other components. Besides, Table \ref{personalized_different_part} also shows that personalizing the normalization layers works better than personalizing the MLP layers and the whole encoder.

\begin{table}[ht]
\renewcommand\arraystretch{1.25}
\caption{
Test accuracy on average for Transformer-based models with different personalized part over 50 clients. 
}
\label{personalized_different_part}
\resizebox{\columnwidth}{!}{
\centering
\begin{tabular}{lllll}
\hline
\multirow{2}{*}{\begin{tabular}[c]{@{}l@{}} Personalized part\end{tabular}} & \multicolumn{2}{c}{CIFAR-10}             & \multicolumn{2}{c}{CIFAR-100}             \\ \cline{2-5} 
    & Pathological        & Dirichlet           & Pathological        & Dirichlet           \\ \hline
Self-attention (FedTP)                                                               & \textbf{90.31±0.26} & \textbf{81.24±0.17} & \textbf{68.05±0.24} & \textbf{46.35±0.29} \\
MLP Layers                                                                           & 88.45±0.14          & 77.36±0.17          & 59.76±0.14          & 35.65±0.16          \\
Normalization Layers                                                                & 89.56±0.45         & 78.15±0.27          & 64.23±0.37          & 41.22±0.39          \\
The Whole Encoder                                                                        & 82.34±0.43          & 73.65±0.52          & 59.79±0.24          & 33.95±0.37          \\ \hline
\end{tabular}
}
\end{table}

\begin{figure*}[htbp]
\centering
\includegraphics[width=0.95\textwidth]{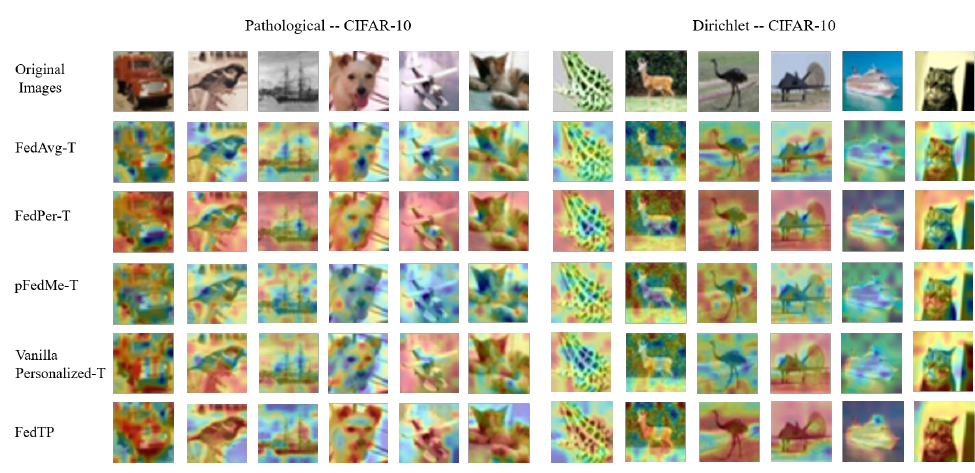} 
\caption{Attention maps of FedTP and other Transformer-based variants implemented on CIFAR-10 over 50 clients.}
\label{attention_10}
\end{figure*}

\begin{figure*}[htbp]
\centering
\includegraphics[width=0.95\textwidth]{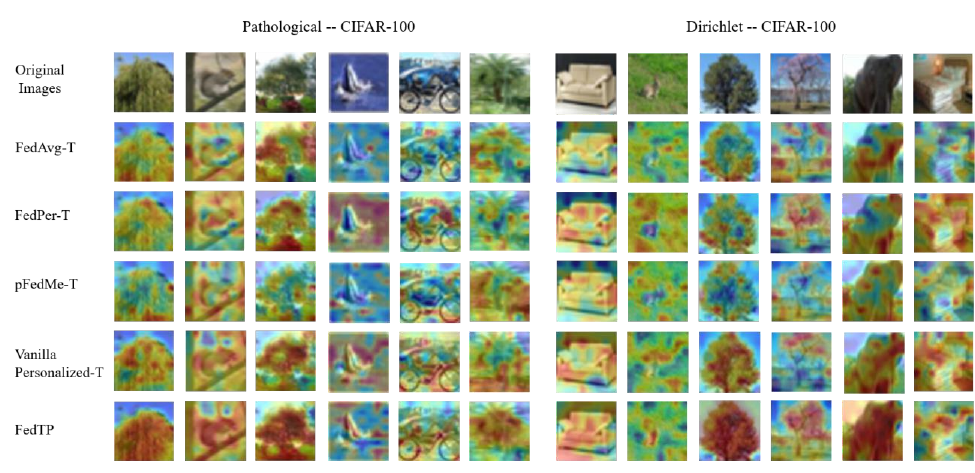} 
\caption{Attention maps of FedTP and other Transformer-based variants implemented on CIFAR-100 over 50 clients.}
\label{attention_100}
\end{figure*}

\emph{5) Visualization of Attention Maps:}
We used Attention Rollout \cite{abnar2020quantifying} to visualize various Transformer-based methods in federated learning. In order to aggregate information across self-attention in each Transformer block, we used the MAX operation instead of the AVG operation proposed in the original paper, and we discarded the least 30$\%$ attention values to filter out low-frequency signals. Fig. \ref{attention_10} and Fig. \ref{attention_100} present visual comparisons of attention maps between FedTP and other Transformer-based variants on CIFAR-10 and CIFAR-100. Both Vanilla Personalized-T and FedTP exhibit client-specific self-attention for good visualization maps. In addition, our FedTP focuses more precisely on the critical parts of testing objects than pFedMe-T and FedAvg-T in many cases. This reflects that FedTP does well in depicting personal attention and again validates our claim.

\emph{6) Extension of FedTP:} 
In this section, we mainly explore the compatibility of FedTP, and our FedTP is compatible with methods with personalized classifier head including FedPer \cite{arivazhagan2019federated}, FedRod \cite{chen2022bridging}, and methods making use of local memory like KNN-Per \cite{marfoq2022personalized}. We inserted those modules into FedTP and evaluated their performance on the CIFAR-10/CIFAR-100 datasets. In our implementation, FedTP+FedPer retains each client's own classification head locally based on FedTP.
FedTP+FedRoD computes the sum of the output of the personalized classification head and the global classification head as the prediction logits.
FedTP+KNN establishes and maintains a local repository in a similar way to Knn-Per which relies on FAISS library \cite{FAISSlibrary}.
The results are shown in Table \ref{ablation_study}. 

\begin{table}[ht]
\renewcommand\arraystretch{1.25}
\caption{
Test accuracy on average for FedTP and its variants over 100 clients. 
}
\label{ablation_study}
\resizebox{1\columnwidth}{!}{
\centering
\begin{tabular}{lllll}
\hline
               & \multicolumn{2}{c}{CIFAR-10}                                              & \multicolumn{2}{c}{CIFAR-100}                                    \\ \cline{2-5} 
\#setting      & \multicolumn{1}{c}{Pathological}    & \multicolumn{1}{c}{Dirichlet}           & \multicolumn{1}{c}{Pathological}    & \multicolumn{1}{c}{Dirichlet}  \\ \hline
FedTP (ours)    & \multicolumn{1}{c}{88.30±0.35} & \multicolumn{1}{c}{79.22±0.29} & \multicolumn{1}{c}{60.90±0.26} & \multicolumn{1}{c}{39.74±0.32} \\
FedTP+FedPer   & 89.42±0.10            & 78.18±0.14                              & 61.78±0.14                     & 32.86±0.19                     \\
FedTP+FedRoD   & 87.66±0.31                     & 78.67±0.37                              & 64.62±0.22           & 42.30±0.26           \\
FedTP+KNN      & 88.84±0.18                     & 80.18±0.12                              & 64.30±0.27                     & 40.70±0.26                     \\ \hline
\end{tabular}
}
\end{table}

\begin{figure}[t]
\centering
\includegraphics[width=0.92\columnwidth]{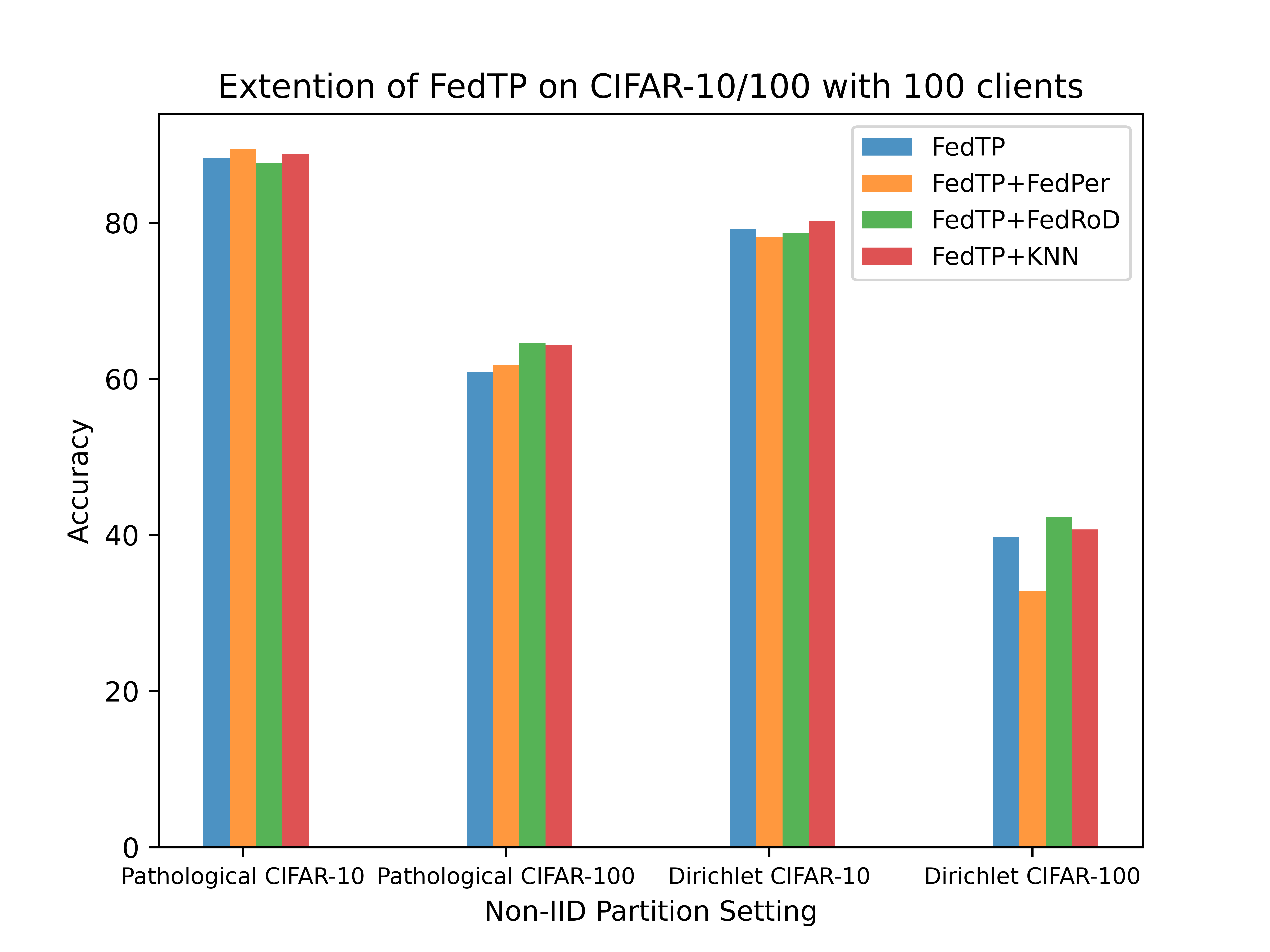} 
\caption{Visualization of accuracy for FedTP Extension Experiments on CIFAR-10 and CIFAR-100 datasets with 100 clients and 5$\%$ sample rate.}
\label{TP_Plus}
\end{figure}

Several discoveries can be found based on Fig. \ref{TP_Plus}. On the one hand, combining local memory to calibrate the classification head can improve the model's performance, especially when the data quantity of each class is small. On the other hand, simply binding a personalized classification head to FedTP may diminish the accuracy when data complies with a Dirichlet distribution. This means a personal classifier may prevent FedTP from learning personalized self-attention when the data distribution is too diverse. Keeping the global head as FedTP+FedRod does will mitigate such problems and improve model performance. To sum up, FedTP can well incorporate former algorithms to help each client better adapt its model to its data heterogeneity.

\begin{figure}[h]
\centering
\includegraphics[width=0.92\columnwidth]{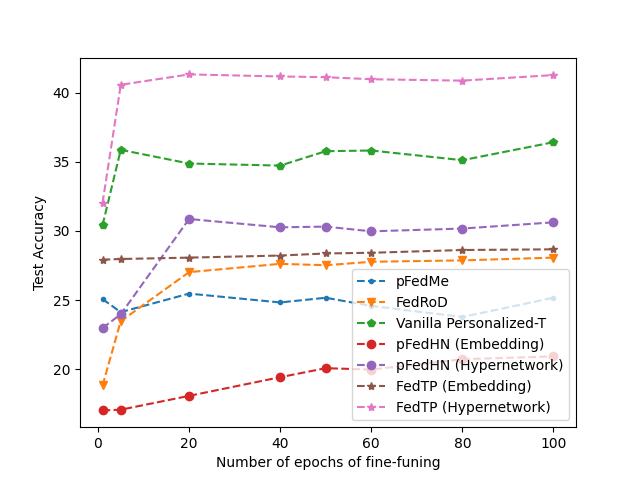}
\caption{Test accuracy of novel clients after fine-tuning the personalized parts of models trained on CIFAR-100.}
\label{fine_tune}
\end{figure}

\emph{7) Generalization to Novel Clients:}
Similar to the settings in \cite{shamsian2021personalized}, we tested our method's generalization ability against pFedMe, pFedHN, FedRod and Vanilla Personalized-T on CIFAR-100 with the Dirichlet setting. Here, we sampled $20\%$ percent of the clients as novel clients whose data is unseen during training, and only the personalized parameters of these models were fine-tuned for novel clients. Specifically, FedRod tuned the personalized parameters in the last classification layer and pFedMe learned all parameters to get each client's personalized model.
Vanilla Personalized-T tuned clients' personalized projection matrices of self-attention layer. With regards to pFedHN and FedTP, the personalized parameters can choose between the client-wise embedding vectors with dimension 32 and the whole hypernetwork. The results are shown in Fig. \ref{fine_tune}, we added ``Embedding'' and ``Hypernetwork'' after pFedHN and FedTP to distinguish between the two patterns. We can easily see that FedTP (Hypernetwork) is more robust and well adapts to novel clients in just a few epochs.

\emph{8) Analysis of Hypernetworks:} 
To analyze the effects of hypernetworks, we compared FedTP with Vanilla Personalized-T, which restores each client's projection parameters $W_i$ locally without using hypernetworks. Table \ref{transformer-based} shows that FedTP has a prominent lead over Vanilla Personalized-T, indicating that hypernetworks indeed play an important role in FedTP. We further notice that even if hypernetworks only produce the parameters of the self-attention layer, it is still good at encoding client-specific information into client embeddings $z_{i}$. The hypernetworks can simultaneously map the client embeddings $z_{i}$  into a manifold parameterized by hypernetwork parameters $\varphi$. 

\begin{figure}[ht]
\centering
\includegraphics[width=0.92\columnwidth]{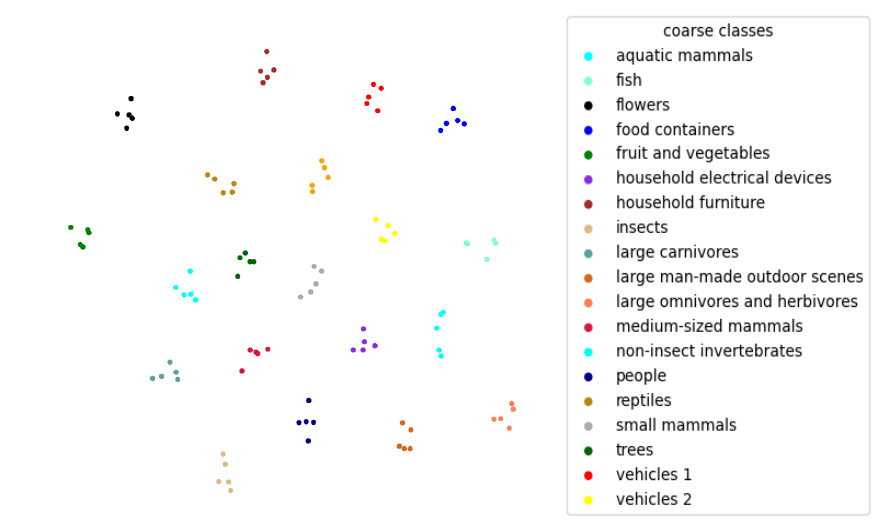} 
\caption{t-SNE visualization of learned client embedding by FedTP on CIFAR-100 dataset.}
\label{t-sne}
\end{figure}

Then we visualize the learned client embeddings by projecting them onto a two-dimensional plane using the t-SNE algorithm \cite{van2008visualizing}. For convenience, we exploited ``coarse'' and ``fine'' labels of CIFAR-100 and split data in a special method similar to pFedHN \cite{shamsian2021personalized}. In detail, we assigned each coarse class to five clients and then split the corresponding fine classes uniformly among those chosen clients. With this extra partition method, we trained FedTP and then visualized client embeddings after training. Fig. \ref{t-sne} shows that the learned personal embeddings of those clients with the same coarse labels are clustered together and they are mapped far away from those with different coarse labels. This result supports our claim that the hypernetwork is a highly effective way of encoding personalized information into their client embeddings $z_i$. 

\subsection{Ablation Study}
\begin{table*}[t]
\centering
\caption{The results of FedTP and the benchmark methods on the image datasets over 100 clients with different $\alpha$ of Dirichlet setting.}
\label{label_distribution}
\renewcommand\arraystretch{1.25}
\resizebox{0.98\textwidth}{!}{
\begin{tabular}{lcccccccccc}
\hline
         & \multicolumn{5}{c}{CIFAR-10}                                   & \multicolumn{5}{c}{CIFAR-100}                                  \\ \cline{2-11} 
$\#\alpha$         & 0.1        & 0.3        & 0.5        & 0.7        & 0.9        & 0.1        & 0.3        & 0.5        & 0.7        & 0.9        \\ \hline
FedAvg-T & 40.99±6.20 & 59.23±1.93 & 63.69±1.33 & 65.29±1.12 & 65.82±0.82 & 32.72±0.81 & 34.89±0.45 & 36.25±1.79 & 36.99±0.44 & 37.51±0.33 \\
FedPer-T & 87.45±0.14 & 77.7±0.14  & 72.44±0.22 & 70.11±0.21 & 71.13±0.14 & 40.92±0.23 & 29.58±0.14 & 27.02±0.11 & 27.12±0.09 & 25.29±0.13 \\
pFedHN   & 83.07±1.07 & 68.36±0.86 & 71.42±0.62 & 68.19±0.81 & 67.62±0.75 & 41.37±0.50 & 29.87±0.69 & 34.55±0.72 & 34.17±0.65 & 33.75±0.58 \\
FedBN    & 84.93±0.53 & 75.41±0.37 & 71.79±0.63 & 69.57±0.56 & 68.70±0.47 & 33.+1±0.04 & 28.70±0.46 & 26.07±0.45 & 26.13±0.35 & 25.08±0.39 \\
FedTP    & \textbf{87.67±0.15} & \textbf{80.27±0.28} & \textbf{75.75±0.26} & \textbf{73.16±0.25} & \textbf{72.12±0.27} & \textbf{48.27±0.26} & \textbf{41.98±0.22} & \textbf{38.83±0.23} & \textbf{38.06±0.31} & \textbf{38.06±0.22} \\ \hline
\end{tabular}
}
\end{table*}

\begin{table*}[ht]
\centering
\renewcommand\arraystretch{1.25}
\caption{The test accuracy of FedTP and FedAvg-T on image datasets over 50 clients with different numbers of self-attention blocks in ViT.}
\label{depth_table}
\resizebox{0.9\textwidth}{!}
{
\begin{tabular}{ccccccccc}
\hline
& \multicolumn{4}{c}{FedAvg-T}                                                            & \multicolumn{4}{c}{FedTP}                                                             \\ \cline{2-9} Block number
                              & \multicolumn{2}{c}{CIFAR-10}              & \multicolumn{2}{c}{CIFAR-100}             & \multicolumn{2}{c}{CIFAR-10}              & \multicolumn{2}{c}{CIFAR-100}             \\ \cline{2-9} 
                              & Pathological             & Dirichlet           & Pathological             & Dirichlet           & Pathological             & Dirichlet           & Pathological             & Dirichlet           \\ \hline
1                             & 36.40±5.48          & 49.38±2.60          & 23.05±1.03          & 27.02±0.44          & 85.15±0.17          & 72.56±0.24          & 59.02±0.30          & 34.97±0.23          \\
2                             & 46.58±4.11          & 55.82±2.31          & 27.58±1.04          & 32.92±0.51          & 88.42±0.17          & 77.04±0.23          & 64.40±0.22          & 41.20±0.27          \\
4                             & 50.21±4.18          & 60.19±1.75          & 31.65±0.85          & 36.23±0.37          & 89.58±0.12          & 78.86±0.26          & 66.35±0.21          & 43.90±0.26          \\
6                             & 49.75±4.12          & 61.34±1.48          & 31.74±0.73          & 36.15±0.32          & 89.95±0.15          & 80.00±0.34          & 66.51±0.35          & 43.69±0.17          \\
8                             & \textbf{50.42±4.22} & 61.85±1.52          & \textbf{34.02±0.88} & \textbf{38.64±0.22} & \textbf{90.31±0.26} & 81.24±0.17          & 68.05±0.24          & \textbf{46.35±0.29} \\
10                            & 46.73±4.43          & \textbf{62.17±1.57} & 33.46±0.82          & 37.52±0.21          & 89.98±0.26          & \textbf{82.02±0.21} & \textbf{69.14±0.44} & 45.42±0.26          \\ \hline
\end{tabular}
}
\end{table*}

\begin{figure*}[t]
\centering
\includegraphics[width=0.9\textwidth]{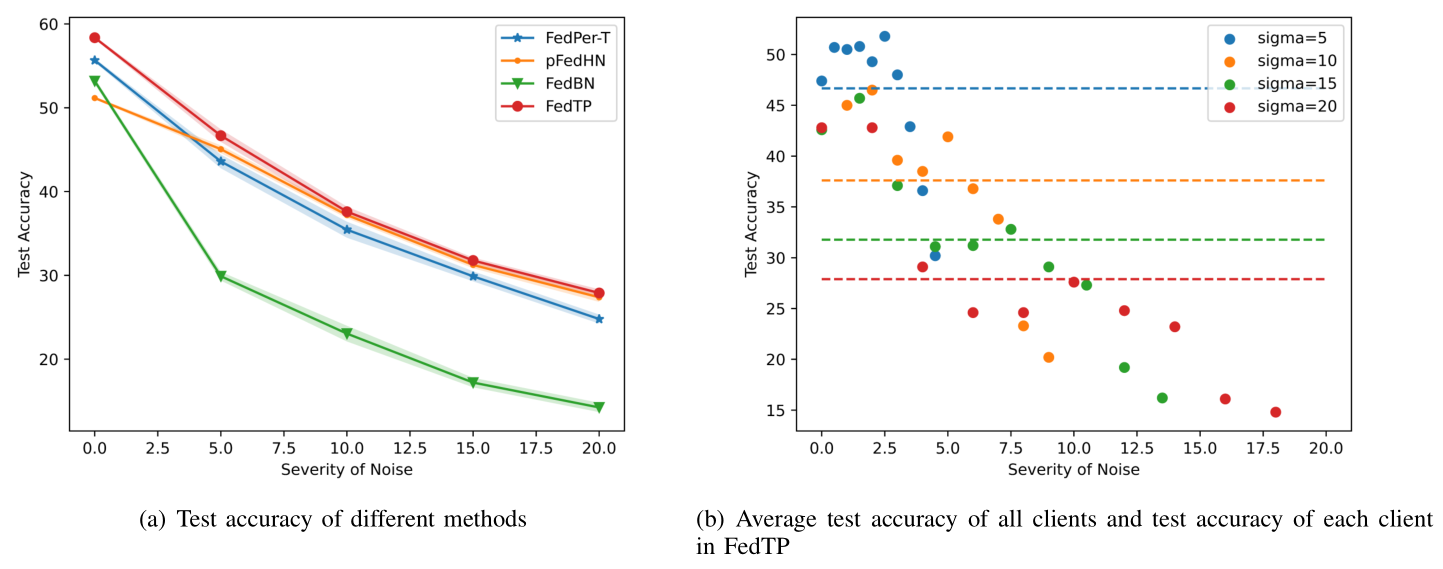} 
\caption{(a) The test accuracy of FedTP and other benchmark methods on CIFAR-10 with different levels of noise; (b) We use different colors to present the different severity of noise, and the points denote the test accuracy of different clients in those different settings while the dotted line presenting the average test accuracy of all clients.}
\label{noise}
\end{figure*}

\emph{1) Effect of Heterogeneity in Label Distribution:}
Data heterogeneity is the key problem that Personalized federated learning aims to solve. We have shown that FedTP outperforms various benchmark methods under several settings. Here, we concentrated on the label distribution heterogeneity. In previous experiments, we tested model performance with this heterogeneity by sampling class in each client following Dirichlet distribution with $\alpha$=0.3. Now we explore more complete cases with $\alpha \in$  $\{0.1, 0.3, 0.5, 0.7, 0.9\}$ for CIFAR-10 and CIFAR-100 datasets. The level of data heterogeneity is higher with a smaller $\alpha$. We set FedAvg-T as baseline and then compared FedTP with various personalized algorithms including FedPer-T, pFedHN \cite{shamsian2021personalized}, FedBN \cite{li2020fedbn}.

From the results in Table \ref{label_distribution}, it can be easily found that with a higher degree of data heterogeneity, performance for FedAvg-T decreases while performance for methods with personalized modules increases. Among these methods, FedTP outperforms them in every case and is much more robust. Hence, FedTP could well overcome label distribution heterogeneity in a wide range. 
As $\alpha$ increases, some personalized federated learning algorithms may fail to make full use of heterogeneity in each client and even falls behind FedAvg-T in accuracy. FedTP, however, is still able to work well. This indicates that FedTP could better mine and exploit heterogeneity among clients even if heterogeneity is not so prominent.

\emph{2) Effect of Heterogeneity in Noise-based Feature Imbalance:}
Generally, noise is another significant factor that may cause data heterogeneity. In the real world, even though each client's data distribution is similar, they may still suffer from different levels of noise and this will lead to feature imbalance among clients. Therefore, we explored the effect of such imbalance with a new partition method. For each client, we add increasing level of random Gaussian noise with mean $\mu=0$ and each client's standard deviation is derived by $\sigma_i = \frac{\sigma_{M}}{N-1} * i$, where $i \in \{0,1,\cdots, N-1\}$. In detail, we let the client number $N$ be 10 with $50\%$ participation rate and conduct a series of experiments with $\sigma_{M} \in \{0, 5, 10, 15, 20\}$. We compared FedTP with various personalized federated methods. Fig. \ref{noise}(a) demonstrates that our FedTP leads to other methods in all different cases. This means FedTP could deal with client-specific noise well. Furthermore, Fig. \ref{noise}(b) shows the average test accuracy of all clients and the test accuracy of each client in FedTP for different levels of noise.

\emph{3) Impact of the Self-Attention Block Number:}
Here we examined the impact of self-attention block number with its value $L \in \{1, 2, 4, 6, 8, 10\}$. Table \ref{depth_table} shows results for FedTP with different block numbers. As can be seen, accumulating attention blocks will help the model develop a better ability to catch its data heterogeneity and improve model behavior to a certain extent. According to this table's results, we choose 8 as our FedTP's default attention block number for CIFAR-10 and CIFAR-100. 

\begin{figure}[h]
\centering
\includegraphics[width=0.88\columnwidth]{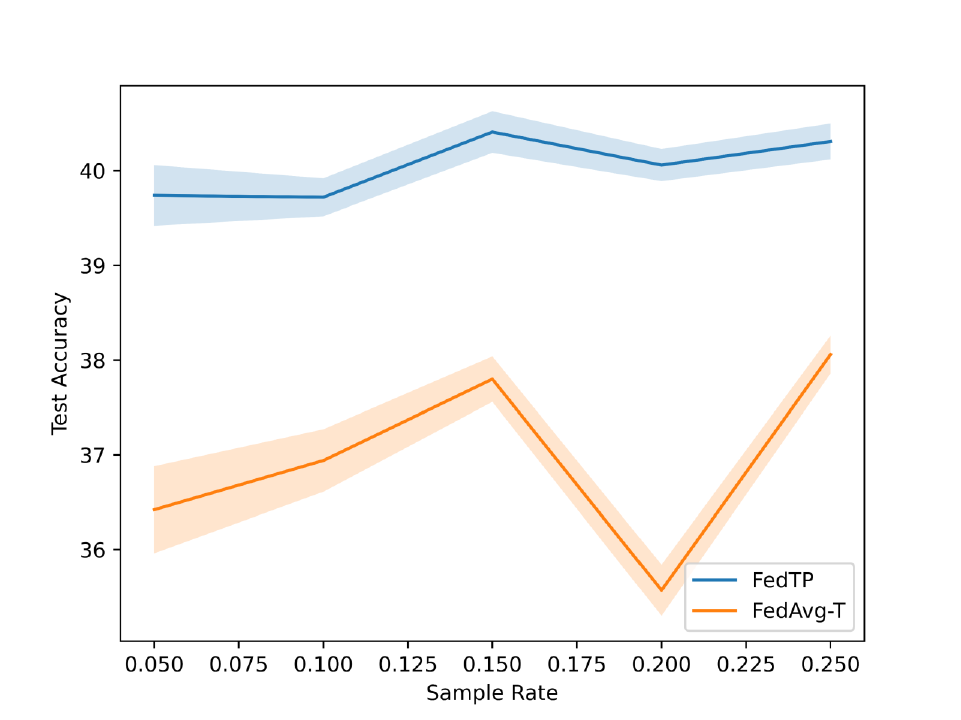} 
\caption{The test accuracy of FedTP and FedAvg-T of Dirichlet setting on CIFAR-100 over 100 clients with different client sample rates.}
\label{sample_rate}
\end{figure}

\emph{4) Impact of Client Sample Rate:}
To explore how the number of participated client impacts model performance, we implemented experiments with sample rate $s\in \{0.05, 0.1, 0.15, 0.2, 0.25\}$. Fig. \ref{sample_rate} clearly illustrates that the performance of FedAvg-T is obviously affected by the sampling rate while our FedTP is relatively more stable. This phenomenon also reflects the robustness of FedTP. 

\section{Conclusions}
We have investigated the impact of self-attention under the federated learning framework and have revealed that FedAvg actually would degrade the performance of self-attention in non-IID scenarios. 
To address this issue, we designed a novel Transformer-based federated learning framework called FedTP that learns personalized self-attention for each client while aggregating the other parameters among the clients. Instead of using the vanilla personalization mechanism that maintains the self-attention layers of each client locally, we proposed the \emph{learn-to-personalize} mechanism to further encourage cooperation among clients and to increase the scalability and generalization of FedTP. 
Specifically, the \emph{learn-to-personalize} is realized by learning a hypernetwork on the server that outputs the personalized projection matrices of self-attention layers to generate client-wise queries, keys and values. Moreover, we also provided the generalization bound for FedTP with the \emph{learn-to-personalize} mechanism. 

Comprehensive experiments have verified that FedTP with the \emph{learn-to-personalize} mechanism delivers great performance in non-IID scenarios and outperforms state-of-the-art methods in personalized federated learning. Since FedTP learns a central hypernetwork over all clients, the model adapts better to novel clients
, which confirms its better generalization to novel clients with the \emph{learn-to-personalize} mechanism. When the local datasets suffer from different levels of noise, the performance of FedTP is better than other benchmarks in all cases, which indicates that FedTP is more robust. 
Notably, FedTP enjoys good compatibility with many advanced personalized federated learning methods, i.e., we can simply combine FedTP with these methods, including FedPer, FedRod and KNN-Per, etc to further enhance the model performance. It can be seen that these combined models have achieved better performance according to the experimental results. The primary limitation of FedTP is its slow training speed compared to CNNs-based federated learning models, a common issue of Transformer-based methods. A more computation-efficient and communication-efficient FedTP is under consideration for our future work.

\bibliographystyle{ieeetr}
\bibliography{reference}

\begin{thebibliography}{10}

\bibitem{mcmahan2017communication}
B.~McMahan, E.~Moore, D.~Ramage, S.~Hampson, and B.~A. y~Arcas,
  ``Communication-efficient learning of deep networks from decentralized
  data,'' in {\em Artificial intelligence and statistics}, pp.~1273--1282,
  PMLR, 2017.

\bibitem{sattler2019robust}
F.~Sattler, S.~Wiedemann, K.-R. M{\"u}ller, and W.~Samek, ``Robust and
  communication-efficient federated learning from non-iid data,'' {\em IEEE
  transactions on neural networks and learning systems}, vol.~31, no.~9,
  pp.~3400--3413, 2019.

\bibitem{sattler2020clustered}
F.~Sattler, K.-R. M{\"u}ller, and W.~Samek, ``Clustered federated learning:
  Model-agnostic distributed multitask optimization under privacy
  constraints,'' {\em IEEE transactions on neural networks and learning
  systems}, vol.~32, no.~8, pp.~3710--3722, 2020.

\bibitem{shi2020consensus}
Y.~Shi, C.-T. Lin, Y.-C. Chang, W.~Ding, Y.~Shi, and X.~Yao, ``Consensus
  learning for distributed fuzzy neural network in big data environment,'' {\em
  IEEE Transactions on Emerging Topics in Computational Intelligence}, vol.~5,
  no.~1, pp.~29--41, 2020.

\bibitem{zhang2020hierarchical}
L.~Zhang, Y.~Shi, Y.-C. Chang, and C.-T. Lin, ``Hierarchical fuzzy neural
  networks with privacy preservation for heterogeneous big data,'' {\em IEEE
  Transactions on Fuzzy Systems}, vol.~29, no.~1, pp.~46--58, 2020.

\bibitem{shi2021distributed}
Y.~Shi, L.~Zhang, Z.~Cao, M.~Tanveer, and C.-T. Lin, ``Distributed
  semisupervised fuzzy regression with interpolation consistency
  regularization,'' {\em IEEE Transactions on Fuzzy Systems}, vol.~30, no.~8,
  pp.~3125--3137, 2021.

\bibitem{geirhos2018imagenet}
R.~Geirhos, P.~Rubisch, C.~Michaelis, M.~Bethge, F.~A. Wichmann, and
  W.~Brendel, ``Imagenet-trained cnns are biased towards texture; increasing
  shape bias improves accuracy and robustness,'' {\em arXiv preprint
  arXiv:1811.12231}, 2018.

\bibitem{vaswani2017attention}
A.~Vaswani, N.~Shazeer, N.~Parmar, J.~Uszkoreit, L.~Jones, A.~N. Gomez,
  {\L}.~Kaiser, and I.~Polosukhin, ``Attention is all you need,'' {\em Advances
  in neural information processing systems}, vol.~30, 2017.

\bibitem{ramachandran2019stand}
P.~Ramachandran, N.~Parmar, A.~Vaswani, I.~Bello, A.~Levskaya, and J.~Shlens,
  ``Stand-alone self-attention in vision models,'' {\em Advances in Neural
  Information Processing Systems}, vol.~32, 2019.

\bibitem{bhojanapalli2021understanding}
S.~Bhojanapalli, A.~Chakrabarti, D.~Glasner, D.~Li, T.~Unterthiner, and
  A.~Veit, ``Understanding robustness of transformers for image
  classification,'' in {\em Proceedings of the IEEE/CVF International
  Conference on Computer Vision}, pp.~10231--10241, 2021.

\bibitem{qu2022rethinking}
L.~Qu, Y.~Zhou, P.~P. Liang, Y.~Xia, F.~Wang, E.~Adeli, L.~Fei-Fei, and
  D.~Rubin, ``Rethinking architecture design for tackling data heterogeneity in
  federated learning,'' in {\em Proceedings of the IEEE/CVF Conference on
  Computer Vision and Pattern Recognition}, pp.~10061--10071, 2022.

\bibitem{dosovitskiy2020image}
A.~Dosovitskiy, L.~Beyer, A.~Kolesnikov, D.~Weissenborn, X.~Zhai,
  T.~Unterthiner, M.~Dehghani, M.~Minderer, G.~Heigold, S.~Gelly, {\em et~al.},
  ``An image is worth 16x16 words: Transformers for image recognition at
  scale,'' {\em arXiv preprint arXiv:2010.11929}, 2020.

\bibitem{krizhevsky2009learning}
A.~Krizhevsky, G.~Hinton, {\em et~al.}, ``Learning multiple layers of features
  from tiny images,'' 2009.

\bibitem{abnar2020quantifying}
S.~Abnar and W.~Zuidema, ``Quantifying attention flow in transformers,'' {\em
  arXiv preprint arXiv:2005.00928}, 2020.

\bibitem{zhang2022r}
W.~Zhang, F.~Yu, X.~Wang, X.~Zeng, H.~Zhao, Y.~Tian, F.-Y. Wang, L.~Li, and
  Z.~Li, ``R$^2$fed: Resilient reinforcement federated learning for industrial
  applications,'' {\em IEEE Transactions on Industrial Informatics}, 2022.

\bibitem{zhang2022semi}
W.~Zhang, X.~Chen, K.~He, L.~Chen, L.~Xu, X.~Wang, and S.~Yang,
  ``Semi-asynchronous personalized federated learning for short-term
  photovoltaic power forecasting,'' {\em Digital Communications and Networks},
  2022.

\bibitem{zhang2022federated}
L.~Zhang, Y.~Shi, Y.-C. Chang, and C.-T. Lin, ``Federated fuzzy neural network
  with evolutionary rule learning,'' {\em IEEE Transactions on Fuzzy Systems},
  2022.

\bibitem{tan2022towards}
A.~Z. Tan, H.~Yu, L.~Cui, and Q.~Yang, ``Towards personalized federated
  learning,'' {\em IEEE Transactions on Neural Networks and Learning Systems},
  2022.

\bibitem{wang2019federated}
K.~Wang, R.~Mathews, C.~Kiddon, H.~Eichner, F.~Beaufays, and D.~Ramage,
  ``Federated evaluation of on-device personalization,'' {\em arXiv preprint
  arXiv:1910.10252}, 2019.

\bibitem{mansour2020three}
Y.~Mansour, M.~Mohri, J.~Ro, and A.~T. Suresh, ``Three approaches for
  personalization with applications to federated learning,'' {\em arXiv
  preprint arXiv:2002.10619}, 2020.

\bibitem{fallah2020personalized}
A.~Fallah, A.~Mokhtari, and A.~Ozdaglar, ``Personalized federated learning with
  theoretical guarantees: A model-agnostic meta-learning approach,'' {\em
  Advances in Neural Information Processing Systems}, vol.~33, pp.~3557--3568,
  2020.

\bibitem{khodak2019adaptive}
M.~Khodak, M.-F.~F. Balcan, and A.~S. Talwalkar, ``Adaptive gradient-based
  meta-learning methods,'' {\em Advances in Neural Information Processing
  Systems}, vol.~32, 2019.

\bibitem{li2020federated}
T.~Li, A.~K. Sahu, M.~Zaheer, M.~Sanjabi, A.~Talwalkar, and V.~Smith,
  ``Federated optimization in heterogeneous networks,'' {\em Proceedings of
  Machine Learning and Systems}, vol.~2, pp.~429--450, 2020.

\bibitem{t2020personalized}
C.~T~Dinh, N.~Tran, and J.~Nguyen, ``Personalized federated learning with
  moreau envelopes,'' {\em Advances in Neural Information Processing Systems},
  vol.~33, pp.~21394--21405, 2020.

\bibitem{li2021ditto}
T.~Li, S.~Hu, A.~Beirami, and V.~Smith, ``Ditto: Fair and robust federated
  learning through personalization,'' in {\em International Conference on
  Machine Learning}, pp.~6357--6368, PMLR, 2021.

\bibitem{mendieta2022local}
M.~Mendieta, T.~Yang, P.~Wang, M.~Lee, Z.~Ding, and C.~Chen, ``Local learning
  matters: Rethinking data heterogeneity in federated learning,'' in {\em
  Proceedings of the IEEE/CVF Conference on Computer Vision and Pattern
  Recognition}, pp.~8397--8406, 2022.

\bibitem{hanzely2020federated}
F.~Hanzely and P.~Richt{\'a}rik, ``Federated learning of a mixture of global
  and local models,'' {\em arXiv preprint arXiv:2002.05516}, 2020.

\bibitem{liang2020think}
P.~P. Liang, T.~Liu, L.~Ziyin, N.~B. Allen, R.~P. Auerbach, D.~Brent,
  R.~Salakhutdinov, and L.-P. Morency, ``Think locally, act globally: Federated
  learning with local and global representations,'' {\em arXiv preprint
  arXiv:2001.01523}, 2020.

\bibitem{marfoq2022personalized}
O.~Marfoq, G.~Neglia, R.~Vidal, and L.~Kameni, ``Personalized federated
  learning through local memorization,'' in {\em International Conference on
  Machine Learning}, pp.~15070--15092, PMLR, 2022.

\bibitem{li2019fedmd}
D.~Li and J.~Wang, ``Fedmd: Heterogenous federated learning via model
  distillation,'' {\em arXiv preprint arXiv:1910.03581}, 2019.

\bibitem{zhu2021data}
Z.~Zhu, J.~Hong, and J.~Zhou, ``Data-free knowledge distillation for
  heterogeneous federated learning,'' in {\em International Conference on
  Machine Learning}, pp.~12878--12889, PMLR, 2021.

\bibitem{liu2021pfa}
B.~Liu, Y.~Guo, and X.~Chen, ``Pfa: Privacy-preserving federated adaptation for
  effective model personalization,'' in {\em Proceedings of the Web Conference
  2021}, pp.~923--934, 2021.

\bibitem{huang2021personalized}
Y.~Huang, L.~Chu, Z.~Zhou, L.~Wang, J.~Liu, J.~Pei, and Y.~Zhang,
  ``Personalized cross-silo federated learning on non-iid data.,'' in {\em
  AAAI}, pp.~7865--7873, 2021.

\bibitem{arivazhagan2019federated}
M.~G. Arivazhagan, V.~Aggarwal, A.~K. Singh, and S.~Choudhary, ``Federated
  learning with personalization layers,'' {\em arXiv preprint
  arXiv:1912.00818}, 2019.

\bibitem{collins2021exploiting}
L.~Collins, H.~Hassani, A.~Mokhtari, and S.~Shakkottai, ``Exploiting shared
  representations for personalized federated learning,'' in {\em International
  Conference on Machine Learning}, pp.~2089--2099, PMLR, 2021.

\bibitem{li2020fedbn}
X.~Li, M.~JIANG, X.~Zhang, M.~Kamp, and Q.~Dou, ``Fedbn: Federated learning on
  non-iid features via local batch normalization,'' in {\em International
  Conference on Learning Representations}, 2020.

\bibitem{achituve2021personalized}
I.~Achituve, A.~Shamsian, A.~Navon, G.~Chechik, and E.~Fetaya, ``Personalized
  federated learning with gaussian processes,'' {\em Advances in Neural
  Information Processing Systems}, vol.~34, pp.~8392--8406, 2021.

\bibitem{li2022data}
Z.~Li, Y.~He, H.~Yu, J.~Kang, X.~Li, Z.~Xu, and D.~Niyato, ``Data
  heterogeneity-robust federated learning via group client selection in
  industrial iot,'' {\em IEEE Internet of Things Journal}, 2022.

\bibitem{shaik2022fedstack}
T.~Shaik, X.~Tao, N.~Higgins, R.~Gururajan, Y.~Li, X.~Zhou, and U.~R. Acharya,
  ``Fedstack: Personalized activity monitoring using stacked federated
  learning,'' {\em Knowledge-Based Systems}, vol.~257, p.~109929, 2022.

\bibitem{zhang2022federated2}
Z.~Zhang, Y.~Jiang, Y.~Shi, Y.~Shi, and W.~Chen, ``Federated reinforcement
  learning for real-time electric vehicle charging and discharging control,''
  in {\em 2022 IEEE Globecom Workshops (GC Wkshps)}, pp.~1717--1722, IEEE,
  2022.

\bibitem{park2021federated}
S.~Park, G.~Kim, J.~Kim, B.~Kim, and J.~C. Ye, ``Federated split vision
  transformer for covid-19cxr diagnosis using task-agnostic training,'' {\em
  arXiv preprint arXiv:2111.01338}, 2021.

\bibitem{ha2016hypernetworks}
D.~Ha, A.~Dai, and Q.~V. Le, ``Hypernetworks,'' {\em arXiv preprint
  arXiv:1609.09106}, 2016.

\bibitem{shamsian2021personalized}
A.~Shamsian, A.~Navon, E.~Fetaya, and G.~Chechik, ``Personalized federated
  learning using hypernetworks,'' in {\em International Conference on Machine
  Learning}, pp.~9489--9502, PMLR, 2021.

\bibitem{ma2022layer}
X.~Ma, J.~Zhang, S.~Guo, and W.~Xu, ``Layer-wised model aggregation for
  personalized federated learning,'' in {\em Proceedings of the IEEE/CVF
  Conference on Computer Vision and Pattern Recognition}, pp.~10092--10101,
  2022.

\bibitem{chen2022bridging}
H.-Y. Chen and W.-L. Chao, ``On bridging generic and personalized federated
  learning,'' {\em International Conference on Learning Representations}, 2022.

\bibitem{caldas2018leaf}
S.~Caldas, S.~M.~K. Duddu, P.~Wu, T.~Li, J.~Kone{\v{c}}n{\`y}, H.~B. McMahan,
  V.~Smith, and A.~Talwalkar, ``Leaf: A benchmark for federated settings,''
  {\em arXiv preprint arXiv:1812.01097}, 2018.

\bibitem{reddi2020adaptive}
S.~Reddi, Z.~Charles, M.~Zaheer, Z.~Garrett, K.~Rush, J.~Kone{\v{c}}n{\`y},
  S.~Kumar, and H.~B. McMahan, ``Adaptive federated optimization,'' {\em arXiv
  preprint arXiv:2003.00295}, 2020.

\bibitem{ConvNet}
Y.~LeCun, L.~Bottou, Y.~Bengio, and P.~Haffner, ``Gradient-based learning
  applied to document recognition,'' {\em Proceedings of the IEEE}, vol.~86,
  no.~11, pp.~2278--2324, 1998.

\bibitem{marfoq2021federated}
O.~Marfoq, G.~Neglia, A.~Bellet, L.~Kameni, and R.~Vidal, ``Federated
  multi-task learning under a mixture of distributions,'' {\em Advances in
  Neural Information Processing Systems}, vol.~34, pp.~15434--15447, 2021.

\bibitem{2019pytorch}
A.~Paszke, S.~Gross, F.~Massa, A.~Lerer, J.~Bradbury, G.~Chanan, T.~Killeen,
  Z.~Lin, N.~Gimelshein, L.~Antiga, {\em et~al.}, ``Pytorch: An imperative
  style, high-performance deep learning library,'' {\em Advances in neural
  information processing systems}, vol.~32, 2019.

\bibitem{FAISSlibrary}
J.~Johnson, M.~Douze, and H.~J{\'{e}}gou, ``Billion-scale similarity search
  with gpus,'' {\em CoRR}, vol.~abs/1702.08734, 2017.

\bibitem{van2008visualizing}
L.~Van~der Maaten and G.~Hinton, ``Visualizing data using t-sne.,'' {\em
  Journal of machine learning research}, vol.~9, no.~11, 2008.

\end{thebibliography}

\vfill
\end{document}